\documentclass[runningheads]{llncs}

\usepackage{eccv}

\usepackage{eccvabbrv}

\usepackage{graphicx}
\usepackage{booktabs}

\usepackage[accsupp]{axessibility}  

\usepackage{amsmath}
\usepackage{amssymb}
\usepackage{arydshln}
\usepackage{booktabs}
\usepackage{comment}
\usepackage{makecell}
\usepackage{multirow}
\usepackage{multicol}
\usepackage{paralist}
\usepackage{pifont}
\usepackage{tikz}
\usepackage{pgfplots}
\usepackage{wrapfig}
\usepackage{enumitem}
\usetikzlibrary{shapes,arrows,positioning}
\usepackage{xcolor,colortbl}
\usepackage{tabularx}
\usepackage{adjustbox}
\usepackage{dashrule}
\usepackage[utf8]{inputenc}
\usepackage[titletoc]{appendix}
\usepackage{titletoc}
\usepackage{nicematrix}
\usepackage{xcolor}

\definecolor{Gray}{gray}{0.93}
\definecolor{uclagold}{rgb}{1.0, 0.7, 0.0}
\definecolor{airforceblue}{rgb}{0.36, 0.54, 0.66}
\definecolor{rosegold}{rgb}{0.72, 0.43, 0.47}
\definecolor{pastelbrown}{rgb}{0.51, 0.41, 0.33}
\definecolor{isabelline}{rgb}{0.96, 0.94, 0.93}
\definecolor{macaroniandcheese}{rgb}{0.98, 0.89, 0.83}
\definecolor{wildblueyonder}{rgb}{0.85, 0.89, 0.95}
\definecolor{mediumtaupe}{rgb}{0.4, 0.3, 0.28}
\definecolor{bluegray}{rgb}{0.4, 0.6, 0.8}
\definecolor{celestialblue}{rgb}{0.29, 0.59, 0.82}
\definecolor{darkorange}{rgb}{1.0, 0.55, 0.0}
\definecolor{cadmiumred}{rgb}{0.89, 0.0, 0.13}
\definecolor{magnolia}{rgb}{0.97, 0.96, 1.0}
\definecolor{pastelblue}{rgb}{0.68, 0.78, 0.81}
\definecolor{persiangreen}{rgb}{0.0, 0.65, 0.58}
\definecolor{steelblue}{rgb}{0.27, 0.51, 0.71}
\definecolor{bluebell}{rgb}{0.64, 0.64, 0.82}
\definecolor{dimgray}{rgb}{0.41, 0.41, 0.41}
\definecolor{splashedwhite}{rgb}{1.0, 0.99, 1.0}
\definecolor{lavendergray}{rgb}{0.77, 0.76, 0.82}
\definecolor{lightgray}{rgb}{0.83, 0.83, 0.83}
\definecolor{lavendermist}{rgb}{0.9, 0.9, 0.98}
\definecolor{lightgreen}{HTML}{f8fcf4}
\definecolor{lightblue}{HTML}{dfebf7}
\definecolor{zeroshot}{rgb}{0.9, 0.9, 0.9}
\definecolor{fourshot}{rgb}{0.8, 0.9, 0.8}
\definecolor{eightshot}{rgb}{0.8, 0.8, 0.9}
\definecolor{sixteenshot}{rgb}{0.9, 0.8, 0.8}
\usetikzlibrary{patterns}

\usepackage{hyperref}

\usepackage{orcidlink}

\usepackage[capitalize]{cleveref}
\crefname{section}{Sec.}{Secs.}
\Crefname{section}{Section}{Sections}
\Crefname{table}{Table}{Tables}
\crefname{table}{Tab.}{Tabs.}

\newcommand{\MLMabbr}{MLLM}
\newcommand{\modelname}{MM1}

\newcommand{\modelnamesmall}{MM1-3B}

\newcommand{\modelnameme}{MM1-7B}
\newcommand{\modelnamelarge}{MM1-30B}
\newcommand{\modelnamesmallchat}{MM1-3B-Chat}
\newcommand{\modelnamechat}{MM1-7B-Chat}
\newcommand{\modelnamelargechat}{MM1-30B-Chat}

\let\oldtimes\times
\renewcommand{\times}{\!\oldtimes\!}

\renewcommand \thepart{}
\renewcommand \partname{}

\begin{document}




\title{MM1: Methods, Analysis \& Insights from Multimodal LLM Pre-training}



\author{Brandon McKinzie$^\circ$ \and
Zhe Gan$^\circ$ \and
Jean-Philippe Fauconnier$^\star$ \and \\
Sam Dodge$^\star$ \and
Bowen Zhang$^\star$ \and
Philipp Dufter$^\star$ \and
Dhruti Shah$^\star$ \and 
Xianzhi Du$^\star$ \and \\
Futang Peng \and
Floris Weers \and
Anton Belyi \and
Haotian Zhang \and 
Karanjeet Singh \and \\
Doug Kang \and
Ankur Jain \and 
Hongyu Hè \and
Max Schwarzer \and
Tom Gunter \and  \\
Xiang Kong \and
Aonan Zhang \and
Jianyu Wang \and
Chong Wang \and
Nan Du \and
Tao Lei \and \\
Sam Wiseman \and 
Guoli Yin \and 
Mark Lee \and
Zirui Wang \and
Ruoming Pang \and \\
Peter Grasch$^\star$ \and 
Alexander Toshev$^\dagger$ \and
Yinfei Yang$^\dagger$}

\authorrunning{B. McKinzie et al.}

\institute{Apple\\
\email{bmckinzie@apple.com}, \email{zhe.gan@apple.com}\\
$^\circ$First authors; $^\star$Core authors; $^\dagger$Senior authors
}

\maketitle
\begin{abstract}
    In this work, we discuss building performant Multimodal Large Language Models (MLLMs). In particular, we study the importance of various architecture components and data choices. Through careful and comprehensive ablations of the image encoder, the vision language connector, and various pre-training data choices, we identified several crucial design lessons. For example, we demonstrate that for large-scale multimodal pre-training using a careful mix of image-caption, interleaved image-text, and text-only data is crucial for achieving state-of-the-art (SOTA) few-shot results across multiple benchmarks, compared to other published multimodal pre-training results. Further, we show that the image encoder together with image resolution and the image token count has substantial impact, while the vision-language connector design is of comparatively negligible importance. By scaling up the presented recipe, we build \textbf{MM1}, a family of multimodal models, including both dense variants up to 30B and mixture-of-experts (MoE) variants up to 64B, that are SOTA in pre-training metrics and achieve competitive performance after supervised fine-tuning on a range of established multimodal benchmarks. Thanks to large-scale pre-training, MM1 enjoys appealing properties such as enhanced in-context learning, and multi-image reasoning, enabling few-shot chain-of-thought prompting.
\end{abstract}
\section{Introduction}
\label{sec:intro}

In recent years, the research community has achieved impressive progress in language modeling and image understanding. Thanks to the availability of large-scale image-text data and compute at scale, we have seen the emergence of highly performant Large Language Models (LLMs)~\cite{brown2020language,devlin2018bert,shoeybi2019megatron,raffel2020exploring,wei2021finetuned,rae2021scaling,thoppilan2022lamda,chowdhery2023palm,zhang2022opt,chung2022scaling,touvron2023llama,bommasani2021opportunities} and Vision Foundation Models~\cite{radford2021learning,oquab2023dinov2,he2022masked} that have become the \emph{de-facto} standard for the majority of language and image understanding problems. Given the above developments, an area of multimodal foundation models has emerged that marries the above advances into a single model achieving superior capabilities. In particular, Multimodal Large Language Models (MLLMs) are large-scale foundation models that consume image and text data and produce text~\cite{li2019visualbert,lu2019vilbert,tsimpoukelli2021multimodal,driess2023palm}. After the rise of LLMs, \MLMabbr{}s are emerging as the next frontier in foundation models. 

\begin{figure}[t!]
\centering
{
\renewcommand{\arrayrulewidth}{0.4mm}
\renewcommand\arraystretch{1.5}
\setlength{\fboxsep}{0pt}
\setlength{\tabcolsep}{3pt}
\setlength{\fboxrule}{0.9pt}
\includegraphics[width=0.97\textwidth]{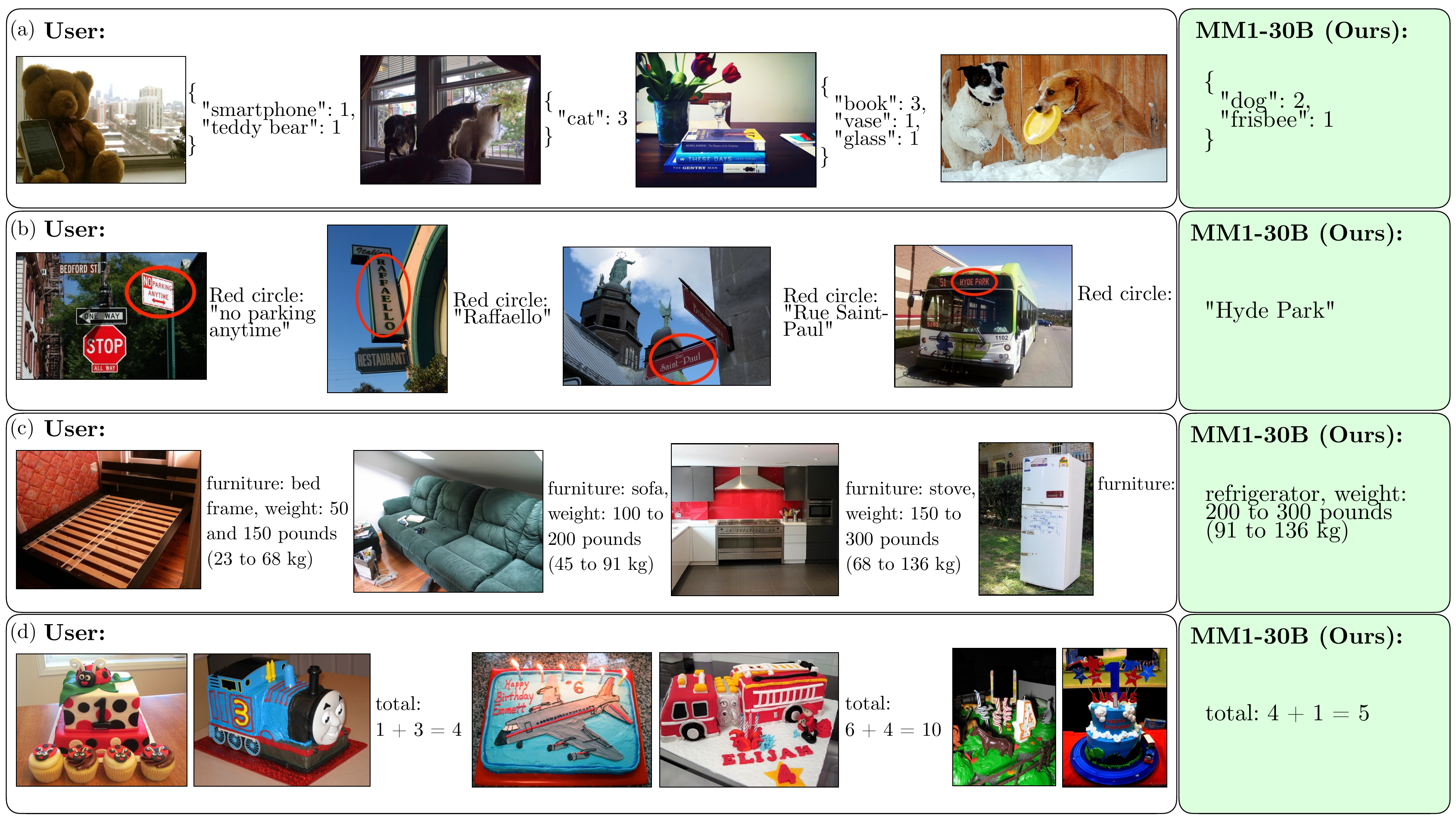}
}
\caption{MM1 can perform in-context predictions thanks to its large-scale multimodal pre-training. This allows MM1 to (a) count objects and follow custom formatting, \mbox{(b) refer} to parts of the images and perform OCR, (c) demonstrate common-sense and word knowledge about everyday objects, and (d) perform basic math functions. Images are from the COCO 2014 validation set \cite{cocodataset}.}
    \label{fig:qualitative_example_mm1_pt}
    \vspace{-2mm}
\end{figure}

When it comes to transparency, existing \MLMabbr{}s fall into two categories: closed models~\cite{achiam2023gpt,team2023gemini} and open models~\cite{flamingo,open-flamingo,peng2023kosmos2,bai2023qwen,liu2023llava}. In the former category, the models might be available for use, but little to nothing is known about the data, model architecture, and training details. In the latter category, the model parameters might be released together with a detailed description of data, model, and training configurations, thus allowing the community to build upon. However, most of the works, both open and closed, release close to nothing about the process they have undergone to arrive at their algorithmic design choices, especially regarding multimodal pre-training.

\begin{figure}[t!]
\centering
{
\renewcommand{\arrayrulewidth}{0.4mm}
\renewcommand\arraystretch{1.5}
\setlength{\fboxsep}{0pt}
\setlength{\tabcolsep}{3pt}
\setlength{\fboxrule}{0.9pt}
\includegraphics[width=0.97\textwidth]{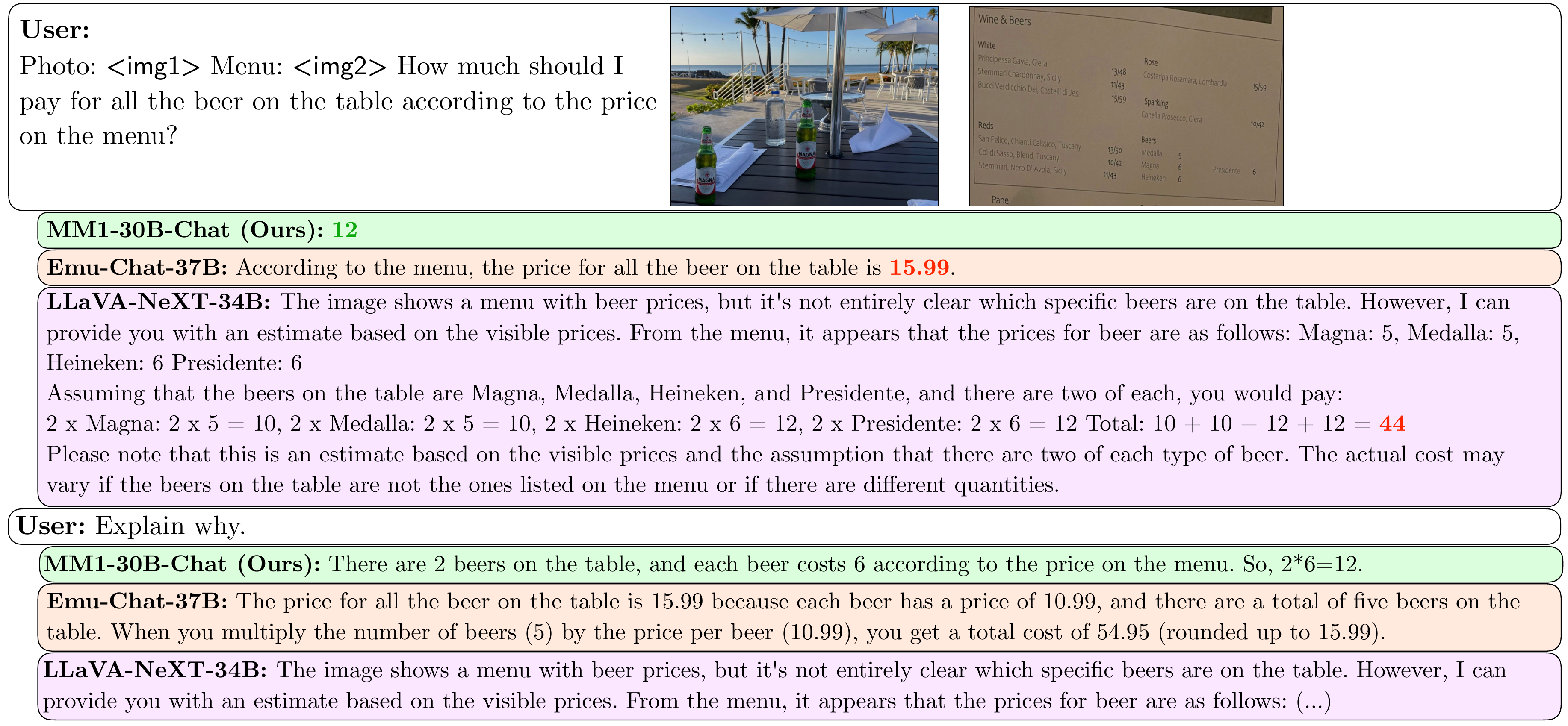}
}
\caption{MM1 can follow instructions and reason across images. Example and images from VILA~\cite{lin2023vila}; VILA answers correctly when prompted with chain-of-thought.}
    \label{fig:qualitative_example_mm1_sft}
    \vspace{-2mm}
\end{figure}

To further research in this area, we believe it is imperative to distill principles and lessons of how to build such models that might outlive concrete component implementations. Thus, in this paper, we document the \MLMabbr{} building process and attempt to formulate design lessons, that we hope are of use to the community.

In particular, our contributions are as follows. First, we perform ablations at small scale across (1) model architecture decisions and (2) pre-training data choices. We identify several interesting trends. On the modeling side, we see that design aspects are in the following order of importance: image resolution, visual encoder loss and capacity, and visual encoder pre-training data. Surprisingly, though, we find little evidence that architectural decisions of how visual data is fed into the LLM matter.

Further, we use three different types of multimodal pre-training data: image-caption, interleaved image-text, and text-only data. We see that when it comes to few-shot and text-only performance, interleaved and text-only training data is of paramount importance, while for zero-shot performance, caption data matters most. We demonstrate that these trends hold after Supervised Fine-Tuning (SFT), both on the evaluations used in the pre-training as well as on further benchmarks. This shows that capabilities and modeling decisions discovered during pre-training are retained after fine-tuning.

Finally, we scale up our model by using larger LLMs, from 3B, 7B, to 30B, and by exploring mixture-of-experts (MoE) models, from 3B with 64 experts to 7B with 32 experts. This leads to a family of performant models, that outperforms most of the relevant works to the best of our knowledge. In particular, the pre-trained model \modelname{} is SOTA, performing better than Emu2~\cite{sun2023generative}, Flamingo~\cite{flamingo}, and IDEFICS~\cite{idefics} on captioning and visual question answering (VQA) tasks in few-shot settings, both in small and large size regimes. The final models, after SFT, achieve competitive performance across 12 established multimodal benchmarks. 

Thanks to large-scale multimodal pre-training, as shown in Figures~\ref{fig:qualitative_example_mm1_pt} and ~\ref{fig:qualitative_example_mm1_sft}, \modelname{} enjoys appealing properties such as in-context predictions, multi-image and chain-of-thought reasoning. MM1 also enables strong few-shot learning capability after instruction tuning.
These strong results demonstrate that the presented recipe for building \MLMabbr{}s translates the design principles to a competitive model at scale. We hope that these presented insights will remain relevant, even as specific modeling components and data sources evolve.
\vspace{-2mm}
\section{Related Work}
\label{sec:related_works}
\vspace{-2mm}
The type of \MLMabbr{}s concerned in this work build upon a strong pre-trained autoregressive LLM that consumes both text and visual tokens, the latter obtained via an image encoder~\cite{bai2023qwen, kosmos-1, peng2023kosmos2,chen2023pali,driess2023palm,blip-2,llava}. Our approach is based on a decoder-only architecture, akin to Kosmos-1~\cite{kosmos-1}.

Recent research has increasingly focused on visual instruction tuning on top of the pre-trained LLM~\cite{li2023multimodal}. Prominent examples include LLaVA(-1.5/NeXT)~\cite{llava,liu2023improved,liu2024llavanext}, MiniGPT-4~\cite{zhu2023minigpt}, mPLUG-Owl(-2/Doc)~\cite{ye2023mplug,ye2023mplug_b,ye2023mplug_c}, Otter~\cite{li2023otter,li2023mimic}, InstructBLIP~\cite{instruct-blip}, Honeybee~\cite{cha2023honeybee}, SPHINX(-X)~\cite{lin2023sphinx,gao2024sphinx}, to name a few. There is also a rich body of literature on constructing instruction-tuning data~\cite{gong2023multimodal,zhao2023svit,li2023m,chen2023sharegpt4v,wang2023see}, enabling MLLMs for referring and grounding~\cite{peng2023kosmos2,chen2023shikra,you2023ferret,wang2023visionllm,lai2023lisa,zhang2023llava}, image generation and editing~\cite{koh2023generating,sun2023generative,fu2023guiding}.

The body of work that focuses on thorough ablations, in particular also on the pre-training side, is relatively sparse. VILA \cite{lin2023vila} focuses on studying various components of multimodal pre-training, but falls short of providing optimization details or detailed pre-training evaluations. Emu2 \cite{sun2023generative}, on the other side, provides details regarding pre-training optimization parameters and base model results. However, they do not provide ablations that justify the various component decisions. IDEFICS \cite{obelics} is another work that provides details regarding large-scale multimodal pre-training. However, their focus is primarily on closely replicating the closed-source Flamingo \cite{flamingo} model.

In contrast to these previous works, we aim to provide details regarding all components of our pre-training strategy, from hyperparameters to data to architecture. We also provide results for our base pre-trained models to help differentiate the impact of multimodal pre-training \emph{vs.} instruction tuning. Furthermore, we provide extensive ablations on the precise impacts of decisions regarding visual encoders, vision-language connectors, and pre-training data mixture.
\vspace{-2mm}
\section{Recipe for Building MM1}
\label{sec:methodology_overview}
\vspace{-2mm}
Building  performant \MLMabbr{}s is a highly empirical endeavor. Although the high-level architectural design and training procedure are clear, their concrete form and execution is not. In this work, we present details of the ablations we have performed to arrive at a performant model. We explore three major axes of design decisions:
\begin{itemize}

    \item \textbf{Architecture}: We investigate different pre-trained image encoders and explore varying ways of connecting LLMs with these encoders.
\item \textbf{Data}: We consider different types of data and their relative mixture weights.
    \item \textbf{Training Procedure}: We explore how to train the \MLMabbr{} including the hyperparameters and what parts of the model to train at what stage.
\end{itemize}

\begin{figure}[t!]
    \centering
    \includegraphics[width=0.48\textwidth]{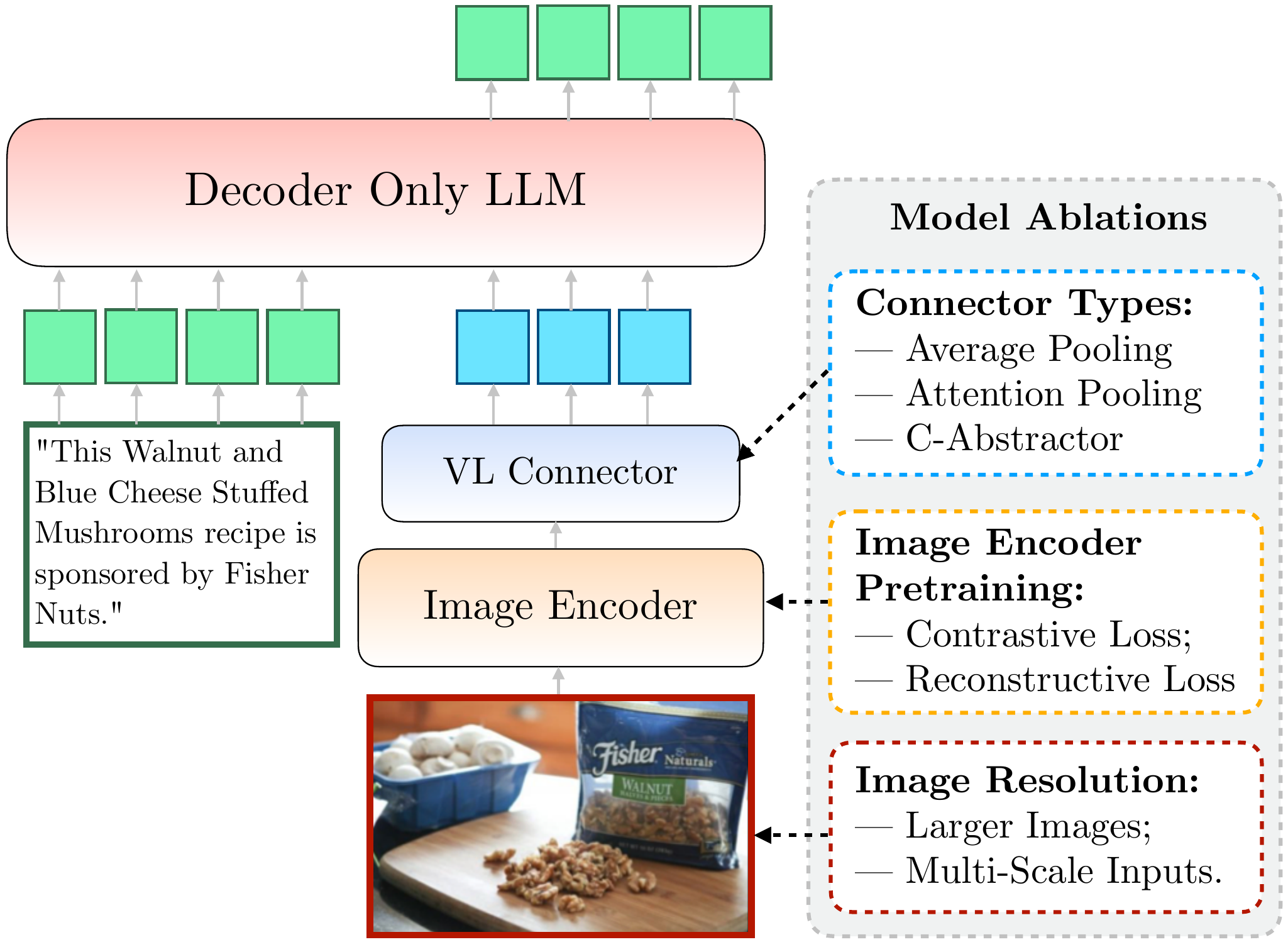}
    \includegraphics[width=0.48\textwidth]{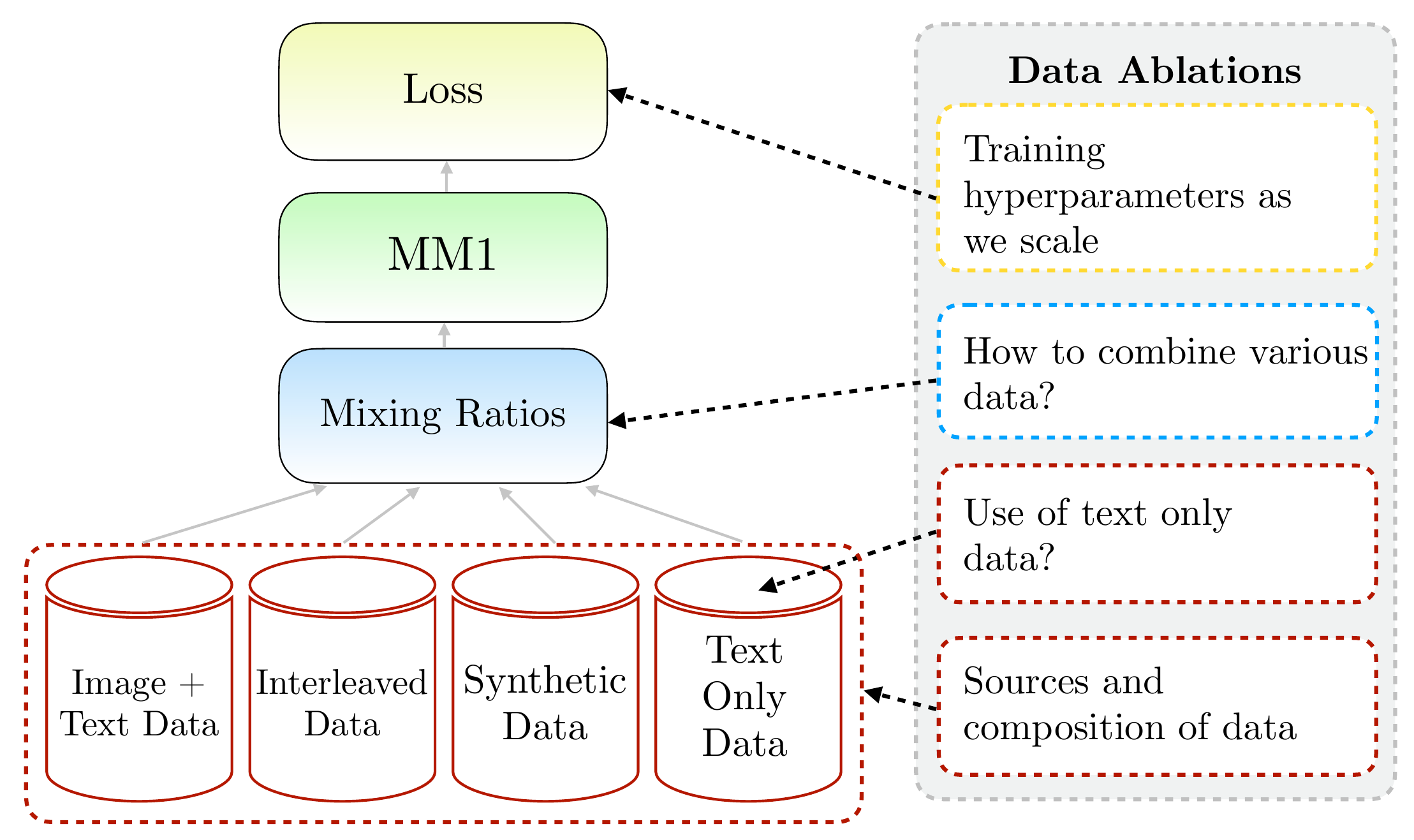}
    \caption{\emph{Left:} Model ablations: what visual encoder to use, how to feed rich visual data, and how to connect the visual representation to the LLM. \emph{Right:} Data ablations: type of data, and their mixture.}
    \label{fig:overview}
    \vspace{-3mm}
\end{figure}

\vspace{-3mm}
\subsection{Empirical Setup for Ablations}
\label{sec:overview_setup}
\vspace{-1mm}
In order to identify what are good choices along each of the above axes, we need an efficient way to assess model performance. As training a large \MLMabbr{} can take substantial resources, we utilize a simplified setup for ablations.

More concretely, we use a smaller base configuration of our model that we ablate from. We modify one component at a time, either an architectural module or a data source, and assess the impact of the design choice for each of these components. This allows us to arrive to the final model-data configuration that we scale up, both in terms of model parameters as well as training time. The base configuration for ablations is as follows:
\begin{itemize}
    \item \textbf{Image Encoder}: A ViT-L/14~\cite{dosovitskiy2020image} model trained with a CLIP loss~\cite{radford2021learning} on DFN-5B~\cite{fang2023data} and VeCap-300M~\cite{veclip}; images of size $336\times 336$.
    \item \textbf{Vision-Language Connector}: C-Abstractor~\cite{cha2023honeybee} with 144 image tokens. 
    \item \textbf{Pre-training Data}: A mix of captioned images (45\%), interleaved image-text documents (45\%), and text-only (10\%) data.
    \item \textbf{Language Model}: A 1.2B transformer decoder-only language model.
\end{itemize}

To evaluate the different design decisions, we use zero-shot and few-shot (4- and 8-shot) performance on a variety of captioning and VQA tasks: COCO Captioning~\cite{chen2015microsoft}, NoCaps~\cite{agrawal2019nocaps}, TextCaps~\cite{sidorov2020textcaps}, VQAv2~\cite{goyal2017making}, TextVQA~\cite{singh2019towards}, VizWiz~\cite{gurari2018vizwiz}, GQA~\cite{hudson2019gqa}, and OK-VQA~\cite{marino2019ok}.
\vspace{-2mm}
\subsection{Model Architecture Ablations}
\label{sec:model_architecture}
\vspace{-1mm}

In this work, we analyze components that enable an LLM to process visual data. Specifically, we investigate (1) how to best pre-train a visual encoder, and (2) how to bridge the visual features to the space of the LLM (see Figure~\ref{fig:overview}, left). 

\begin{table*}[t!]
    \centering
    \resizebox{0.9\textwidth}{!}{%
    \begin{tabular}{cllclccc}
        \toprule
        \multicolumn{5}{c}{\bf Setup} & \multicolumn{3}{c}{\bf Results} \\
        \hline
         & Model & Arch. & Image Res. & Data & 0-shot & 4-shot & 8-shot \\
        \midrule
        \parbox[t]{2mm}{\multirow{3}{*}{\rotatebox[origin=c]{90}{ Recon.}}} & AIM$_{\text{600M}}$ & ViT/600M & \multirow{3}{*}{224} & \multirow{3}{*}{DFN-2B} & 36.6 &  56.6 & 60.7 \\
        & AIM$_{\text{1B}}$ & ViT/1B &  & & 37.9 & 59.5 & 63.3 \\
        & AIM$_{\text{3B}}$ & ViT/3B &  & & 38.9 & 60.9 & 64.9 \\
        \midrule
         \multirow{7}{*}{\rotatebox[origin=c]{90}{Contrastive}} & CLIP$_{\text{DFN+VeCap}}$ & ViT-L & \multirow{3}{*}{224} & DFN-5B$+$VeCap & 36.9 & 58.7 & 62.2 \\
        & CLIP$_{\text{DFN}}$ & ViT-H &  & DFN-5B & 37.5 & 57.0 & 61.4 \\
        & CLIP$_{\text{DFN+VeCap}}$ & ViT-H & & DFN-5B$+$VeCap & 37.5 & 60.0 & 63.6\\
        \cmidrule{2-8}
        & CLIP$_{\text{DFN+VeCap}}$ & ViT-L & \multirow{3}{*}{336} & \multirow{2}{*}{DFN-5B$+$VeCap} & 39.9 & 62.4 & 66.0 \\
        & CLIP$_{\text{DFN+VeCap}}$ & ViT-H &  &  & 40.5 & \bf 62.6 & 66.3 \\
        & CLIP$_{\text{OpenAI}}$ & ViT-L & & ImageText-400M & 39.3 & 62.2 & 66.1 \\
        \cmidrule{2-8}
        & CLIP$_{\text{DFN}}$ & ViT-H & 378 & DFN-5B & \bf 40.9 & 62.5 & \bf 66.4 \\
        \bottomrule
    \end{tabular}}
    \vspace{1mm}
    \caption{\modelname{} pre-training ablation across different image encoders (with 2.9B LLM). Note that the values in the Data column correspond to the data that was used for the initial training of the image encoder itself, not \modelname. Recon.: Reconstructive loss. AIM:~\cite{elnouby2024scalable}; DFN-2/5B:~\cite{fang2023data}; VeCap: VeCap-300M~\cite{veclip}; OpenAI~\cite{radford2021learning}.}
    \label{tab:visual_encoders_ablations}
    \vspace{-6mm}
\end{table*}

\vspace{1mm}
\noindent \textbf{Image Encoder Pre-training.}
Most MLLMs use a CLIP pre-trained image encoder~\cite{llava,liu2023improved,ye2023mplug,instruct-blip}, while recent works also started to explore vision-only self-supervised models, such as DINOv2~\cite{lin2023sphinx,tong2024eyes}, as the image encoder.
Similar to these prior works, we find that the choice of the pre-trained image encoder can substantially impact downstream results both after multimodal pre-training and after instruction tuning. 
Here, we primarily ablate the importance of image resolution and image encoder pre-training objective. Note that unlike the rest of our ablations, here we use a 2.9B LLM (instead of 1.2B) to ensure there is sufficient capacity to utilize some of the larger image encoders. 

\vspace{1mm}
\noindent\textit{Contrastive losses.} When trained on large-scale image-text datasets, the resulting models possess strong semantic understanding of the image data as evidenced by performance on various forms of image classification and retrieval tasks~\cite{radford2021learning}. These results were enabled because of the availability of large-scale image-text data, which can endow a visual encoder with semantic knowledge. More recently, automatically curated large-scale datasets and synthetic captions have led to even stronger encoders~\cite{fang2023data,veclip}.

\vspace{1mm}
\noindent\textit{Reconstructive Losses.} When it comes to dense prediction, CLIP-style models struggle to attain the same strong performance~\cite{rao2022denseclip,wang2023sclip,ranasinghe2023perceptual}. This property can be problematic for MLLMs, as many of the tasks such as VQA and captioning require detailed image understanding. Hence, we also consider image encoders learned using reconstructive losses, as such losses explicitly capture all parts of an image. In particular, we utilize AIM~\cite{elnouby2024scalable}, which has shown that a carefully designed autoregressive reconstructive loss on image data alone scales well.

\textbf{Encoder Lesson: Image resolution has the highest impact, followed by model size and training data composition.} As we can see in Table~\ref{tab:visual_encoders_ablations}, increasing image resolution from $224$ to $336$ results in approx. $3\%$ boost in all metrics across all architectures. Increasing the model size from ViT-L to ViT-H, a doubling in parameters, results in a modest performance increase of usually less than $1\%$. Finally, adding VeCap-300M~\cite{veclip}, a dataset of synthetic captions, yields more than $1\%$ boost in few-shot scenarios.

When it comes to model type, the results are less conclusive. Contrastive methods tend to result in higher performance than reconstructive. In particular, encoders based on ViT-L of 300M parameters result in $0.3\%$ to $1.5\%$ performance gain compared to $\textrm{AIM}_{\text{600M}}$ of comparable size (only 20 of the 24 AIM model layers are used at inference). This lesson is, nevertheless, inconclusive for the potential of AIM as it has been trained on less than half the data. Similarly, the widely used open sourced OpenAI model~\cite{radford2021learning} perform on-par with our model of comparable capacity but trained on DFN+VeCap data mixture.

\begin{figure}[t!]
    \centering
    \includegraphics[width=0.98\textwidth]{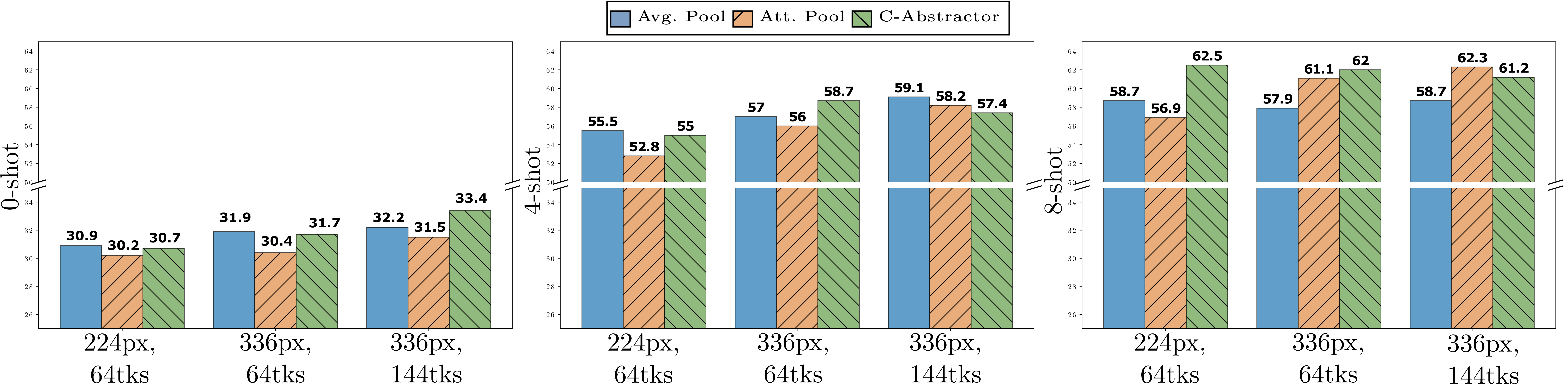}
    \caption{0-shot, 4-shot, and 8-shot ablations across different visual-language connectors for two image resolutions, and two image token sizes.}
    \label{fig:visual_language_bridge_ablations}
    \vspace{-5mm}
\end{figure}

\vspace{1mm}
\noindent \textbf{Vision-Language Connector and Image Resolution.}
 The goal of this component is to translate the visual representation to the space of the LLM. As image encoders are ViTs, their output is either a single embedding, or a set of grid-arranged embeddings corresponding to the input image patches. Therefore, the spatial arrangement of the image tokens needs to be converted to the sequential one of the LLM. At the same time, the actual image token representations are to be mapped to the word embedding space.

While doing so, there are two conflicting requirements. On the one side, we would like to capture as much detail from the image as possible, fulfilled by increasing the number of image token embeddings. On the other side, especially in the case of multi-image input, having a large number of input tokens per image is computationally challenging.

We consider using $64$ or $144$ tokens to represent the image, as well as two different image resolutions, $224$ and $336$. Further, we consider the following architectural options:

\textit{Average Pooling.} Following \cite{sun2023generative}, we apply $n\times n$ average pooling on the output of the ViT image encoder, followed by a linear projection ($n\in\{8, 12\}$).

\textit{Attention Pooling.} Motivated by the fact that image token representations are in a different space than the LLM input embeddings, attention pooling using $k$ learnable queries, is a natural approach. By varying $k$ one can vary the number of inputs from a single image that are fed into the LLM (we use $k\in\{64, 144\})$.

\textit{Convolutional Mapping.} More recently, Honeybee~\cite{cha2023honeybee} has studied the above questions and proposed the C-Abstractor module. It is implemented as a ResNet~\cite{he2016deep} block that preserves local information while through adaptive pooling can change the number of image tokens. 

\textbf{VL Connector Lesson: Number of visual tokens and image resolution matters most, while the type of VL connector has little effect.} The results shown in Figure~\ref{fig:visual_language_bridge_ablations} demonstrate that both zero- and few-shot performance increases as we increase the number of visual tokens or/and image resolution. However, contrary to what has been reported in the literature~\cite{cha2023honeybee}, different architectural designs do not appear to conclusively produce stronger models. After instruction tuning, all three architectures achieve very similar results at the 336px and 144 token setting. (See Appendix
Figure~\ref{fig:sft_ablations} for fine-tuning results.)

\vspace{-2mm}
\subsection{Pre-training Data Ablation}
\label{sec:data}
\vspace{-1mm}
 
Large-scale and task-appropriate data is of paramount importance in training performant models. 
Typically, models are trained in two stages, pre-training and instruction tuning. In the former stage web-scale data is used while in the latter stage task-specific curated data is utilized. In the following, we focus on the pre-training stage and elaborate our data choices (see Figure~\ref{fig:overview}, right).

\begin{table}[h]
\vspace{-2mm}
\setlength{\tabcolsep}{5pt}
\small
    \centering
    \resizebox{\linewidth}{!}{  
    \begin{tabular}{l  l  l }
        \toprule
        \bf{Data Type} & \bf{Sources} & \bf{Size} \\
        \hline
         \multirow{2}{*}{Captioned Images} & CC3M~\cite{sharma2018conceptual}, CC12M~\cite{changpinyo2021conceptual}, HQIPT-204M~\cite{ranasinghe2023perceptual}, & \multirow{2}{*}{2B image-text pairs} \\
         & COYO~\cite{kakaobrain2022coyo-700m}, Web Image-Text-1B (Internal)  & \\
         \cdashline{1-3}
         \rule{-4pt}{1.05\normalbaselineskip}
         Captioned Images (Synthetic) & VeCap~\cite{veclip} & 300M image-text pairs \\
         \cdashline{1-3}
         \rule{-4pt}{1.05\normalbaselineskip}
         Interleaved Image-Text& OBELICS~\cite{obelics}, Web Interleaved (Internal) & 600M documents \\
         \cdashline{1-3}
         \rule{-4pt}{1.05\normalbaselineskip}
         \multirow{2}{*}{Text-only}  & Webpages, Code, Social media,  & \multirow{2}{*}{2T tokens} \\
         & Books, Encyclopedic, Math & \\
         \bottomrule
    \end{tabular}
    }
    \vspace{1mm}
    \caption{List of datasets for pre-training multimodal large language models.}
    \label{tab:data_ablate_sources}
    \vspace{-7mm}
\end{table}

Two types of data are commonly used to train MLLMs: captioning data consisting of images with paired text descriptions; and interleaved image-text documents from the web (see Appendix~\ref{sec:app:interleaved} for details). 
Note that captioning data tends to contain relatively short text with high relevance to the image. On the contrary, interleaved data has substantially longer and more diverse text with less relevance, on average, to the surrounding images. Finally, we include text-only data to help preserve the language understanding capabilities of the underlying pre-trained LLM. The full list of datasets is summarized in Table~\ref{tab:data_ablate_sources}.

We use the same model setup for ablations described in Section~\ref{sec:overview_setup}, with the only exception that we train 200k steps here to fully leverage the large-scale data training. 
We also incorporate a set of commonly employed text tasks, referred to as TextCore\footnote{TextCore tasks include ARC~\cite{arc}, PIQA~\cite{piqa}, LAMBADA~\cite{paperno2016lambada}, WinoGrande~\cite{winogrande}, HellaSWAG~\cite{zellers2019hellaswag}, SciQ~\cite{sciq}, TriviaQA~\cite{joshi2017triviaqa}, and WebQS~\cite{webqs}.}, as part of the evaluation to better assess the effects of data mixture.
These lead to the following lessons:

\begin{figure}[tb]
     \centering
     \begin{subfigure}[b]{0.48\textwidth}
         \centering
         \definecolor{bblue}{HTML}{4F81BD}
\definecolor{rred}{HTML}{C0504D}
\definecolor{ggreen}{HTML}{9BBB59}
\definecolor{ppurple}{HTML}{9F4C7C}

\definecolor{myorange}{RGB}{252,213,180}
\definecolor{myblue}{RGB}{184,204,228}
\definecolor{myred}{RGB}{230,184,183}
\definecolor{mypurple}{RGB}{204,192,218}
\definecolor{myturquoise}{RGB}{183,222,232}
\definecolor{mybrown}{RGB}{196,189,151}
\definecolor{mygray}{RGB}{217,217,217}
\definecolor{mygreen}{RGB}{216,228,188}
\definecolor{myblue2}{RGB}{141,180,226}

\definecolor{mpl_blue}{HTML}{769FC7}
\definecolor{mpl_orange}{HTML}{F3AA72}
\definecolor{mpl_green}{HTML}{85BB77}
\definecolor{mpl_red}{HTML}{D6756F}
\definecolor{mpl_purple}{HTML}{AF96CD}

\makeatletter 
\pgfdeclarepatternformonly[\LineSpace,\tikz@pattern@color]{my north east lines}{\pgfqpoint{-1pt}{-1pt}}{\pgfqpoint{\LineSpace}{\LineSpace}}{\pgfqpoint{\LineSpace}{\LineSpace}}%
{
    \pgfsetcolor{\tikz@pattern@color} 
    \pgfsetlinewidth{0.4pt}
    \pgfpathmoveto{\pgfqpoint{0pt}{0pt}}
    \pgfpathlineto{\pgfqpoint{\LineSpace + 0.1pt}{\LineSpace + 0.1pt}}
    \pgfusepath{stroke}
}
\pgfdeclarepatternformonly[\LineSpace,\tikz@pattern@color]{my north west lines}{\pgfqpoint{-1pt}{-1pt}}{\pgfqpoint{\LineSpace}{\LineSpace}}{\pgfqpoint{\LineSpace}{\LineSpace}}%
{
    \pgfsetcolor{\tikz@pattern@color} 
    \pgfsetlinewidth{0.4pt}
    \pgfpathmoveto{\pgfqpoint{0pt}{\LineSpace}}
    \pgfpathlineto{\pgfqpoint{\LineSpace + 0.1pt}{-0.1pt}}
    \pgfusepath{stroke}
}
\makeatother 
\newdimen\LineSpace
\tikzset{
    line space/.code={\LineSpace=#1},
    line space=3pt
}

    \usetikzlibrary{
        matrix,
    }
    \pgfplotsset{
        compat=1.3,
    }

\begin{tikzpicture}
    \begin{axis}[
        width  = \textwidth*1.15,
        height = 5cm,
        major x tick style = transparent,
        ybar=0pt,
        bar width=5pt,
        ymajorgrids = true,
        ylabel = {Average Performance},
        symbolic x coords={TextCore,0-shot,4-shot,8-shot},
        x tick label style  = {text width=1.5cm,align=center},
        xtick = data,
        ylabel near ticks,
        xlabel near ticks,
        scaled y ticks = false,
        enlarge x limits=0.25,
        ymin=20,
        ymax=80,
        legend cell align=left,
        legend style={at={(0.5,1.03)},
            anchor=north,legend columns=5,
            font=\tiny,
            inner sep=1pt,
            },
        legend image code/.code={
        \draw [#1] (-0.07cm,-0.1cm) rectangle (0.15cm,0.08cm); },
        nodes near coords,
        every node near coord/.append style={font=\tiny, rotate=90, anchor=west},
        nodes near coords align={vertical},
        yticklabel style = {font=\tiny,xshift=0.5ex},
        xticklabel style = {font=\scriptsize,yshift=0.5ex},
        ylabel style = {font=\footnotesize,yshift=-1.5ex},
        title style = {font=\small\bfseries,yshift=-1.5ex,},
    ]

        \addplot[style={black,fill=mpl_blue,mark=none}]
            coordinates {(TextCore, 49.6) (0-shot,39.3) (4-shot,43.8) (8-shot,45.0)};

        \addplot[style={black,fill=mpl_orange,mark=none,postaction={pattern=my north east lines, line space=8pt}}]
             coordinates {(TextCore, 51.7) (0-shot,35.9) (4-shot,58.0) (8-shot,61.1)};

        \addplot[style={black,fill=mpl_green,mark=none,postaction={pattern=my north west lines, line space=8pt}}]
             coordinates {(TextCore, 52.2) (0-shot,33.4) (4-shot,58.7) (8-shot,62.2)};        

        \addplot[style={black,fill=mpl_red,mark=none,postaction={pattern=vertical lines}}]
             coordinates {(TextCore, 52.0) (0-shot,33.1) (4-shot,58.2) (8-shot,61.9)};

        \addplot[style={black,fill=mpl_purple,mark=none,postaction={pattern=horizontal lines}}]
             coordinates {(TextCore,52.8) (0-shot,25.8) (4-shot,53.6) (8-shot,56.9)};

        \legend{100/0, 66/33, 50/50, 33/66, 0/100}
        
    \end{axis}
\end{tikzpicture}\\[-6ex]
         \caption{Caption/Interleaved Mixing }
        \label{fig:caption_interleaved_mixing}
     \end{subfigure}
     \hfill
     \begin{subfigure}[b]{0.48\textwidth}
         \centering
         \definecolor{bblue}{HTML}{4F81BD}
\definecolor{rred}{HTML}{C0504D}
\definecolor{ggreen}{HTML}{9BBB59}
\definecolor{ppurple}{HTML}{9F4C7C}

\definecolor{myorange}{RGB}{252,213,180}
\definecolor{myblue}{RGB}{184,204,228}
\definecolor{myred}{RGB}{230,184,183}
\definecolor{mypurple}{RGB}{204,192,218}
\definecolor{myturquoise}{RGB}{183,222,232}
\definecolor{mybrown}{RGB}{196,189,151}
\definecolor{mygray}{RGB}{217,217,217}
\definecolor{mygreen}{RGB}{216,228,188}
\definecolor{myblue2}{RGB}{141,180,226}

\definecolor{mpl_blue}{HTML}{769FC7}
\definecolor{mpl_orange}{HTML}{F3AA72}
\definecolor{mpl_green}{HTML}{85BB77}
\definecolor{mpl_red}{HTML}{D6756F}
\definecolor{mpl_purple}{HTML}{AF96CD}

\newdimen\LineSpace
\tikzset{
    line space/.code={\LineSpace=#1},
    line space=3pt
}

    \usetikzlibrary{
        matrix,
    }
    \pgfplotsset{
        compat=1.3,
    }

\begin{tikzpicture}
    \begin{axis}[
        width  = \textwidth*1.15,
        height = 5cm,
        major x tick style = transparent,
        ybar=0pt,
        bar width=6pt,
        ymajorgrids = true,
        ylabel = {Average Performance},
        symbolic x coords={TextCore,0-shot,4-shot,8-shot},
        x tick label style  = {text width=1.5cm,align=center},
        xtick = data,
        ylabel near ticks,
        xlabel near ticks,
        scaled y ticks = false,
        enlarge x limits=0.25,
        ymin=20,
        ymax=80,
        legend cell align=left,
        legend style={at={(0.5,1.03)},
            anchor=north,legend columns=2,
            font=\tiny,
            inner sep=1pt,
            },
        legend image code/.code={
        \draw [#1] (-0.07cm,-0.1cm) rectangle (0.15cm,0.08cm); },
        nodes near coords,
        every node near coord/.append style={font=\tiny, rotate=90, anchor=west},
        nodes near coords align={vertical},
        yticklabel style = {font=\tiny,xshift=0.5ex},
        xticklabel style = {font=\scriptsize,yshift=0.5ex},
        ylabel style = {font=\footnotesize,yshift=-1.5ex},
        title style = {font=\small\bfseries,yshift=-1.5ex,},
    ]

        \addplot[style={black,fill=mpl_blue,mark=none}]
            coordinates {(TextCore, 49.6) (0-shot,39.3) (4-shot,43.8) (8-shot,45.0)};

        \addplot[style={black,fill=mpl_orange,mark=none,postaction={pattern=my north east lines, line space=8pt}}]
             coordinates {(TextCore, 54.8) (0-shot,35.3) (4-shot,51.4) (8-shot,53.6)};

        \addplot[style={black,fill=mpl_green,mark=none,postaction={pattern=my north west lines, line space=8pt}}]
             coordinates {(TextCore, 52.8) (0-shot,25.8) (4-shot,53.6) (8-shot,56.9)};       

        \addplot[style={black,fill=mpl_red,mark=none,postaction={pattern=vertical lines}}]
             coordinates {(TextCore, 54.5) (0-shot,24.0) (4-shot,51.6) (8-shot,55.3)};

        \legend{Caption, Caption+Text, Interleaved, Interleaved+Text}
        
    \end{axis}
\end{tikzpicture}\\[-6ex]
         \caption{Importance of Text-Only Data }
         \label{fig:text_only_data_importance}
     \end{subfigure}
     \\
     \begin{subfigure}[b]{0.48\textwidth}
         \centering
         \definecolor{bblue}{HTML}{4F81BD}
\definecolor{rred}{HTML}{C0504D}
\definecolor{ggreen}{HTML}{9BBB59}
\definecolor{ppurple}{HTML}{9F4C7C}

\definecolor{myorange}{RGB}{252,213,180}
\definecolor{myblue}{RGB}{184,204,228}
\definecolor{myred}{RGB}{230,184,183}
\definecolor{mypurple}{RGB}{204,192,218}
\definecolor{myturquoise}{RGB}{183,222,232}
\definecolor{mybrown}{RGB}{196,189,151}
\definecolor{mygray}{RGB}{217,217,217}
\definecolor{mygreen}{RGB}{216,228,188}
\definecolor{myblue2}{RGB}{141,180,226}

\definecolor{mpl_blue}{HTML}{769FC7}
\definecolor{mpl_orange}{HTML}{F3AA72}
\definecolor{mpl_green}{HTML}{85BB77}
\definecolor{mpl_red}{HTML}{D6756F}
\definecolor{mpl_purple}{HTML}{AF96CD}

\newdimen\LineSpace
\tikzset{
    line space/.code={\LineSpace=#1},
    line space=3pt
}

    \usetikzlibrary{
        matrix,
    }
    \pgfplotsset{
        compat=1.3,
    }

\begin{tikzpicture}
    \begin{axis}[
        width  = \textwidth*1.15,
        height = 5cm,
        major x tick style = transparent,
        ybar=0pt,
        bar width=6pt,
        ymajorgrids = true,
        ylabel = {Average Performance},
        symbolic x coords={TextCore,0-shot,4-shot,8-shot},
        x tick label style  = {text width=1.5cm,align=center},
        xtick = data,
        ylabel near ticks,
        xlabel near ticks,
        scaled y ticks = false,
        enlarge x limits=0.25,
        ymin=20,
        ymax=80,
        legend cell align=left,
        legend style={at={(0.5,1.03)},
            anchor=north,legend columns=4,
            font=\tiny,
            inner sep=1pt,
            },
        legend image code/.code={
        \draw [#1] (-0.07cm,-0.1cm) rectangle (0.15cm,0.08cm); },
        nodes near coords,
        every node near coord/.append style={font=\tiny, rotate=90, anchor=west},
        nodes near coords align={vertical},
        yticklabel style = {font=\tiny,xshift=0.5ex},
        xticklabel style = {font=\scriptsize,yshift=0.5ex},
        ylabel style = {font=\footnotesize,yshift=-1.5ex},
        title style = {font=\small\bfseries,yshift=-1.5ex,},
    ]

        \addplot[style={black,fill=mpl_blue,mark=none}]
            coordinates {(TextCore, 52.2) (0-shot,33.4) (4-shot,58.7) (8-shot,62.2)};

        \addplot[style={black,fill=mpl_orange,mark=none,postaction={pattern=my north east lines, line space=8pt}}]
             coordinates {(TextCore, 54.0) (0-shot,32.1) (4-shot,58.3) (8-shot,62.7)};

        \addplot[style={black,fill=mpl_green,mark=none,postaction={pattern=my north west lines, line space=8pt}}]
             coordinates {(TextCore, 54.2) (0-shot,32.5) (4-shot,57.9) (8-shot,60.8)};       

        \addplot[style={black,fill=mpl_red,mark=none,postaction={pattern=vertical lines}}]
             coordinates {(TextCore, 54.6) (0-shot,32.1) (4-shot,57.1) (8-shot,61.0)};

        \legend{100/0, 91/9, 86/14, 66/33}
        
    \end{axis}
\end{tikzpicture}\\[-6ex]
         \caption{Image/Text-Only Mixing Ablations}
         \label{fig:image_text_mixing_ablations}
     \end{subfigure}
     \hfill
     \begin{subfigure}[b]{0.48\textwidth}
         \centering
         \definecolor{bblue}{HTML}{4F81BD}
\definecolor{rred}{HTML}{C0504D}
\definecolor{ggreen}{HTML}{9BBB59}
\definecolor{ppurple}{HTML}{9F4C7C}

\definecolor{myorange}{RGB}{252,213,180}
\definecolor{myblue}{RGB}{184,204,228}
\definecolor{myred}{RGB}{230,184,183}
\definecolor{mypurple}{RGB}{204,192,218}
\definecolor{myturquoise}{RGB}{183,222,232}
\definecolor{mybrown}{RGB}{196,189,151}
\definecolor{mygray}{RGB}{217,217,217}
\definecolor{mygreen}{RGB}{216,228,188}
\definecolor{myblue2}{RGB}{141,180,226}

\definecolor{mpl_blue}{HTML}{769FC7}
\definecolor{mpl_orange}{HTML}{F3AA72}
\definecolor{mpl_green}{HTML}{85BB77}
\definecolor{mpl_red}{HTML}{D6756F}
\definecolor{mpl_purple}{HTML}{AF96CD}

\newdimen\LineSpace
\tikzset{
    line space/.code={\LineSpace=#1},
    line space=3pt
}

    \usetikzlibrary{
        matrix,
    }
    \pgfplotsset{
        compat=1.3,
    }

\begin{tikzpicture}
    \begin{axis}[
        width  = \textwidth*1.15,
        height = 5cm,
        major x tick style = transparent,
        ybar=0pt,
        bar width=13pt,
        ymajorgrids = true,
        ylabel = {Average Performance},
        symbolic x coords={TextCore,0-shot,4-shot,8-shot},
        x tick label style  = {text width=1.5cm,align=center},
        xtick = data,
        ylabel near ticks,
        xlabel near ticks,
        scaled y ticks = false,
        enlarge x limits=0.25,
        ymin=20,
        ymax=80,
        legend cell align=left,
        legend style={at={(0.5,1.03)},
            anchor=north,legend columns=4,
            font=\tiny,
            inner sep=1pt,
            },
        legend image code/.code={
        \draw [#1] (-0.07cm,-0.1cm) rectangle (0.15cm,0.08cm); },
        nodes near coords,
        every node near coord/.append style={font=\tiny, rotate=90, anchor=west},
        nodes near coords align={vertical},
        yticklabel style = {font=\tiny,xshift=0.5ex},
        xticklabel style = {font=\scriptsize,yshift=0.5ex},
        ylabel style = {font=\footnotesize,yshift=-1.5ex},
        title style = {font=\small\bfseries,yshift=-1.5ex,},
    ]

        \addplot[style={black,fill=mpl_blue,mark=none}]
             coordinates {(TextCore, 53.9) (0-shot,35.4) (4-shot,55.9) (8-shot,58.7)};

        \addplot[style={black,fill=mpl_orange,mark=none,postaction={pattern=my north east lines, line space=8pt}}]
            coordinates {(TextCore, 54.0) (0-shot,32.1) (4-shot,58.3) (8-shot,62.7)};

        \legend{w/o VeCap, w/ VeCap}
        
    \end{axis}
\end{tikzpicture}\\[-6ex]
         \caption{Impact of VeCap Data}
         \label{fig:impact_of_vecap}
     \end{subfigure}
    \caption{Data Ablations. For each ablation, we present four different metrics: TextCore, 0-shot, 4-shot, and 8-shot. \textbf{(a)} Results with image data where we present five different mixing ratios between interleaved and captioned data. \textbf{(b)} Results with and without text-only data. We mix the text-only data separately with captioned and interleaved data. \textbf{(c)} Results with different mixing ratios between image data~(caption and interleaved) and text-only data. \textbf{(d)} Results with and without including VeCap as part of caption data.}
    \label{fig:data-ablations}
    \vspace{-4mm}
\end{figure}

\vspace{1mm}
\noindent\textbf{Data Lesson 1: Interleaved data is instrumental for few-shot and text-only performance, while captioning data lifts zero-shot performance.} In Figure~\ref{fig:caption_interleaved_mixing}, we present results across different mixes of interleaved and captioned data. Zero-shot performance increases consistently, from $25.8\%$ to $39.3\%$, as we increase the amount of captioned data. At the same time, however, for 4- and 8-shot performance, having at least $50\%$ of the data being interleaved is crucial to maintain over $61\%$ for 8-shot or $58\%$ for 4-shot. Without it, performance drops drastically to $45\%$ and $43.7\%$, respectively.
Since interleaved data naturally contains multiple images and accompanying text which are often inter-related, such data is inherently similar to few-shot test inputs, which aligns well with empirical results. However, due to the nature of common evaluation being heavily tailored to captioning problems (3 out of the 8 benchmarks are captioning), captioning data notably lifts zero-shot performance. Interestingly, the use of interleaved data further boosts performance on these very same captioning benchmarks in few-shot settings.
Similarly, text-only performance benefits from interleaved data, likely as interleaved data contains long-form text as well.

\vspace{1mm}
\noindent\textbf{Data Lesson 2: Text-only data helps with few-shot and text-only performance.} We utilize text-only data as a way to maintain the language understanding capabilities of the model. As seen in Figure~\ref{fig:text_only_data_importance}, combining text-only and captioned data boost few-shot performance. In other words, long text does allow the model to utilize multiple image and text examples as context to perform better question answering and captioning. On the other side, combining text-only with interleaved data leads to a drop in performance, albeit a minor one. In both cases, text-only performance is increased as shown in the boost of TextCore numbers.

\vspace{1mm}
\noindent\textbf{Data Lesson 3: Careful mixture of image and text data can yield optimal multimodal performance and retain strong text performance.} The above lesson leads to the question of how to best combine text-only data to achieve both strong image and language understanding. In Figure~\ref{fig:image_text_mixing_ablations}, we experiment with several mixing ratios between image (caption and interleaved) and text-only data. We see that with caption/interleaved/text ratio 5:5:1, we achieve a good balance of strong multimodal performance while still keeping comparable text-only understanding performance. 

\vspace{1mm}
\noindent\textbf{Data Lesson 4: Synthetic data helps with few-shot learning.} At last, we study the importance of the synthetic caption data, VeCap~\cite{veclip}. It is of higher quality, but relatively small, being only $7\%$ compared to all caption data. As shown in Figure~\ref{fig:impact_of_vecap}, it does give a non-trivial boost in few-shot performance, of $2.4\%$ and $4\%$ absolute.
\section{Final Model and Training Recipe}

We collect the results from the previous ablations to determine the final recipe for \modelname{} multimodal pre-training:
\begin{itemize}

    \item \textbf{Image Encoder}: Motivated by the importance of image resolution, we use a ViT-H~\cite{dosovitskiy2020image} model with $378\times378$ resolution, pre-trained with a CLIP objective on DFN-5B~\cite{fang2023data}. 

    \item \textbf{Vision-Language Connector}: As the number of visual tokens is of highest importance, we use a VL connector with 144 tokens. The actual architecture seems to matter less, we opt for C-Abstractor~\cite{cha2023honeybee}.

    \item \textbf{Data}: In order to maintain both zero- and few-shot performance, we use the following careful mix of 45\% interleaved image-text documents, 45\% image-text pair documents, and 10\% text-only documents. 
\end{itemize}

\begin{wrapfigure}[12]{r}{0.45\textwidth}
{
\addtolength\abovecaptionskip{-20pt}  
\vspace*{-1.0cm} 
    \begin{tikzpicture}

\definecolor{darkgray176}{RGB}{176,176,176}
\definecolor{steelblue31119180}{RGB}{31,119,180}

\begin{axis}[
width=\textwidth*0.45,
height=\textwidth*0.3,
log basis x={10},
log basis y={10},
tick align=outside,
tick pos=left,
x grid style={darkgray176},
xlabel={Num LLM Params},
xmin=568852.930843842, xmax=139636836105.588,
xmode=log,
xtick style={color=black},
xtick={10000000,100000000,1000000000,10000000000},
xticklabels={
  \(\displaystyle {10^{7}}\),
  \(\displaystyle {10^{8}}\),
  \(\displaystyle {10^{9}}\),
  \(\displaystyle {10^{10}}\)
},
y grid style={darkgray176},
ylabel={Peak Learning Rate},
ymin=1.1560524123394e-05, ymax=0.00215970881599622,
ymode=log,
ytick style={color=black},
ytick={1e-06,1e-05,0.0001,0.001,0.01,0.1},
yticklabels={
  \(\displaystyle {10^{-6}}\),
  \(\displaystyle {10^{-5}}\),
  \(\displaystyle {10^{-4}}\),
  \(\displaystyle {10^{-3}}\),
  \(\displaystyle {10^{-2}}\),
  \(\displaystyle {10^{-1}}\)
},
grid=both,
    grid style={line width=.1pt, draw=gray!10},
    major grid style={line width=.2pt,draw=gray!50},
    axis line style={latex-latex},
        yticklabel style = {font=\tiny,xshift=0.5ex},
        xticklabel style = {font=\tiny,yshift=0.5ex},
        ylabel style = {font=\tiny,yshift=-1.5ex},
        xlabel style = {font=\tiny},
        title style = {font=\tiny,xshift=-1.5ex,},
]
\addplot [draw=steelblue31119180, fill=steelblue31119180, mark=*, only marks, opacity=0.2]
table{%
x  y
9000000 0.001
9000000 0.000464158883361278
9000000 0.001
9000000 0.001
9000000 0.000464158883361278
9000000 0.001
9000000 0.001
9000000 0.000464158883361278
9000000 0.000464158883361278
9000000 0.000464158883361278
85000000 0.000215443469003188
85000000 0.000215443469003188
85000000 0.000215443469003188
85000000 0.000215443469003188
85000000 0.000215443469003188
85000000 0.000215443469003188
85000000 0.000215443469003188
85000000 0.000215443469003188
85000000 0.000215443469003188
85000000 0.000215443469003188
302000000 0.000215443469003188
302000000 0.000215443469003188
302000000 0.000215443469003188
302000000 0.000215443469003188
302000000 0.000215443469003188
302000000 0.000215443469003188
302000000 0.000215443469003188
302000000 0.000215443469003188
302000000 0.000215443469003188
302000000 0.000215443469003188
1200000000 0.0001
1200000000 4.64158883361278e-05
1200000000 0.0001
1200000000 0.0001
1200000000 4.64158883361278e-05
1200000000 0.0001
1200000000 0.0001
1200000000 4.64158883361278e-05
1200000000 4.64158883361278e-05
1200000000 0.0001
};
\addplot [ultra thick, steelblue31119180]
table {%
1000000 0.00170274237067197
1258925.41179417 0.00154528152830253
1584893.19246111 0.00140238185343955
1995262.31496887 0.00127269680432724
2511886.43150957 0.00115500436045438
3162277.66016837 0.00104819550747109
3981071.70553495 0.000951263786961238
5011872.33627269 0.000863295812598009
6309573.44480189 0.000783462663316568
7943282.34724276 0.000711012072401782
9999999.99999992 0.000645261338888854
12589254.1179416 0.000585590894481124
15848931.924611 0.000531438465366154
19952623.1496886 0.000482293773917
25118864.3150955 0.000437693730352843
31622776.6016834 0.000397218069050082
39810717.0553492 0.00036048538838489
50118723.3627265 0.000327149556790732
63095734.4480184 0.000296896451165171
79432823.4724269 0.000269440996892007
99999999.9999984 0.000244524481586915
125892541.179415 0.000221912117254059
158489319.246109 0.000201390827881892
199526231.496884 0.000182767241630705
251188643.150953 0.000165865868692324
316227766.016831 0.000150527447651962
398107170.553489 0.000136607444769996
501187233.627261 0.000123974692042447
630957344.480179 0.0001125101512066
794328234.724263 0.000102105792045025
999999999.999975 9.26635744182541e-05
1258925411.79414 8.40945244338439e-05
1584893192.46107 7.63178960465532e-05
1995262314.96883 6.92604101894234e-05
2511886431.50951 6.28555642660936e-05
3162277660.16829 5.70430054977113e-05
3981071705.53486 5.1767962219491e-05
5011872336.27257 4.69807277680376e-05
6309573444.80174 4.26361920961115e-05
7943282347.24256 3.8693416701247e-05
9999999999.99967 3.51152488627821e-05
12589254117.9412 3.18679715522609e-05
15848931924.6106 2.89209857183183e-05
19952623149.6881 2.6246521952221e-05
25118864315.0949 2.3819378817095e-05
31622776601.6826 2.16166853751174e-05
39810717055.3482 1.96176856749703e-05
50118723362.7253 1.78035431687845e-05
63095734448.0168 1.61571632155969e-05
79432823472.425 1.46630319987738e-05
};
\end{axis}

\end{tikzpicture}\\[-6ex]
    \caption{Optimal peak learning rate as a function of model size. The data points represent experiments that achieved close-to-optimal 8-shot performance for their associated model size.}
    \label{fig:lr-opt-fit}
    }
\end{wrapfigure}

In order to improve the model performance, we scale up the LLM size to 3B, 7B, and 30B parameters. We initialize both the image encoder and the underlying LLM decoder weights for \modelname{} from in-house pre-trained models\footnote{The LLM is pre-trained on the text-only data mixture mentioned in Sec.~\ref{sec:data}.}. We then perform multimodal pre-training on the above data mix for 200k steps (approx. 400B tokens). All models are pre-trained entirely unfrozen with sequence length 4096, up to 16 images per sequence at $378\times378$ resolution, with a batch size of 512 sequences. All models are trained using the AXLearn framework.\footnote{\url{https://github.com/apple/axlearn}} 

\vspace{1mm}
\noindent\textbf{Model Scaling.} At this scale it is infeasible to do proper hyperparameter search. Instead, using established scaling characteristics of LLMs~\cite{henighan2020scaling,hoffmann2022training,yang2020feature,mup}, we perform a grid search of learning rate at small scale, 9M, 85M, 302M, and 1.2B, while using the components identified in Sec.~\ref{sec:model_architecture}\footnote{The only exception is image encoder, which we downsize to the CLIP$_{\text{DFN+VeCap}}$ ViT-L with $336\times336$ resolution to reduce compute costs for the grid searches.} to identify optimal learning rate and extrapolate it to larger scale. We use a linear regression in log space to extrapolate from smaller to larger models (see Figure~\ref{fig:lr-opt-fit}),  
resulting in the following prediction of optimal peak learning rate $\eta$ given the number of (non-embedding) parameters $N$:%
\vspace{-1em}
\begin{align}
    \eta = \exp(-0.4214\ln(N) - 0.5535) \label{eq:opt-lr}
\end{align}
Similar to \cite{isik2024scaling}, we found in preliminary experiments that validation loss wasn't strongly correlated with downstream task performance. Therefore, we directly use downstream 8-shot average performance for curve fitting.

For $N=3e^{10}$, this fit predicts $\eta = 2.2e^{-5}$, which is what we use for the final \modelnamelarge{}. We initially performed a similar procedure to determine reasonable values for weight decay, denoted by $\lambda$, but ultimately found that the simple rule of scaling weight decay by peak learning rate as $\lambda = 0.1 \eta$ worked well for all models. All further training details are described in Appendix~\ref{app:training-details}.

\vspace{1mm}
\noindent\textbf{Scaling via Mixture-of-Experts (MoE).} MoE scales the total number of model parameters while keeping the activated parameters constant. It enjoys a larger model capacity without sacrificing inference speed significantly. Recently, MoE has shown promising results in language~\cite{zoph2022stmoe,pmlr-v162-du22c, fedus2022switch,jiang2024mixtral,dai2024deepseekmoe}, multimodal~\cite{mustafa2022multimodal,lin2024moellava} and computer vision~\cite{ruiz2021scaling,komatsuzaki2023sparseupcycle,daxberger2023mobile,Chen_2023_ICCV} tasks.

In experiments, we further explore scaling the dense model by adding more experts in the FFN layers of the language model. Our MoE implementation generally follows GShard~\cite{lepikhin2021gshard} and ST-MoE~\cite{zoph2022stmoe}. Specifically, we design two MoE models, a 3B-MoE using 64 experts that replaces a dense layer with a sparse layer in every-2 layers and a 7B-MoE using 32 experts that replaces a dense layer with a sparse layer in every-4 layers. The 3B-MoE contains 64B parameters in total and the 7B-MoE contains 47B parameters in total. We adopt top-2 gating with a load balance loss term with a $0.01$ coefficient to encourage a better expert load balance and adopt a router z-loss term with a $0.001$ coefficient to stabilize training. To convert a dense model to MoE, we only replace the dense language decoder with an MoE language decoder. The image encoder and the vision-language connector are kept the same. To train an MoE, we adopt the same training hyperparameters that are discovered for the dense backbone\footnote{The dense backbone is defined to be the dense model we use to construct the MoE model.} and identical training settings including training data and training tokens.

\begin{table*}[t!]
\centering
\small
\resizebox{0.92\linewidth}{!}{%
\begin{tabular}{l c ccc cccc}
\toprule 
\multirow{2}{*}{\bf Model} 
    & \multirow{2}{*}{~\bf Shot~} 
    & \multicolumn{3}{c}{\bf Captioning} 
    & \multicolumn{4}{c}{\bf Visual Question Answering} 
\\
\cmidrule(lr){3-5} \cmidrule(lr){6-9}
    & 
    & COCO 
    & NoCaps 
    & TextCaps
    & VQAv2 
    & TextVQA 
    & VizWiz 
    & OKVQA \\ 
    \midrule
\multicolumn{9}{l}{\emph{MM1-3B Model Comparisons}} \\
\midrule
\multirow{2}{*}{Flamingo-3B~\cite{flamingo}}
    & \multicolumn{1}{c}{~0$^\dagger$}  
        & 73.0 
        & --
        & --
        & 49.2 
        & 30.1 
        & 28.9 
        & 41.2 
        \\
& \multicolumn{1}{c}{8}
        & 90.6\cellcolor{isabelline} 
        & --\cellcolor{isabelline}
        & --\cellcolor{isabelline}
        & 55.4\cellcolor{isabelline} 
        & 32.4\cellcolor{isabelline} 
        & 38.4\cellcolor{isabelline} 
        & 44.6\cellcolor{isabelline} 
        \\
\cdashline{1-9}
\rule{-4pt}{1.05\normalbaselineskip}
\multirow{2}{*}{\modelnamesmall{}}
    & \multicolumn{1}{c}{0} 
        & 73.5 
        & 55.6 
        & 63.3 
        & 46.2 
        & 29.4 
        & 15.6 
        & 26.1 
        \\
 & \multicolumn{1}{c}{8}   
        & \textbf{114.6}\cellcolor{isabelline} 
        & \textbf{104.7}\cellcolor{isabelline} 
        & \textbf{88.8}\cellcolor{isabelline} 
        & \textbf{63.6}\cellcolor{isabelline} 
        & \textbf{44.6}\cellcolor{isabelline} 
        & \textbf{46.4}\cellcolor{isabelline} 
        & \textbf{48.4}\cellcolor{isabelline} 
        \\
\midrule
\multicolumn{9}{l}{\emph{MM1-7B Model Comparisons}} \\
\midrule
\multirow{2}{*}{IDEFICS-9B~\cite{obelics}}
    & \multicolumn{1}{c}{~0$^\dagger$} 
        & 46.0* 
        & 36.8 
        & 25.4 
        & 50.9 
        & 25.9 
        & 35.5 
        & 38.4 
        \\
    & \multicolumn{1}{c}{8} 
        & 97.0*\cellcolor{isabelline} 
        & 86.8\cellcolor{isabelline} 
        & 63.2\cellcolor{isabelline} 
        & 56.4\cellcolor{isabelline} 
        & 27.5\cellcolor{isabelline} 
        & 40.4\cellcolor{isabelline} 
        & 47.7\cellcolor{isabelline} 
        \\
\cdashline{1-9}
\rule{-4pt}{1.05\normalbaselineskip}
\multirow{2}{*}{Flamingo-9B~\cite{flamingo}}
    & \multicolumn{1}{c}{~0$^\dagger$}  
        & 79.4 
        & --
        & --
        & 51.8 
        & 31.8 
        & 28.8 
        & 44.7 
        \\
    & \multicolumn{1}{c}{8} 
        & 99.0\cellcolor{isabelline} 
        & --\cellcolor{isabelline}
        & --\cellcolor{isabelline}
        & 58.0\cellcolor{isabelline} 
        & 33.6\cellcolor{isabelline} 
        & 39.4\cellcolor{isabelline} 
        & 50.0\cellcolor{isabelline} 
        \\
\cdashline{1-9}
\rule{-4pt}{1.05\normalbaselineskip}
\multirow{2}{*}{Emu2-14B~\cite{sun2023generative}}
    & \multicolumn{1}{c}{~0$^\dagger$}  
        & -- 
        & --
        & --
        & 52.9 
        & --
        & 34.4 
        & 42.8 
        \\
    & \multicolumn{1}{c}{8} 
        & --\cellcolor{isabelline} 
        & --\cellcolor{isabelline}
        & --\cellcolor{isabelline}
        & 59.0\cellcolor{isabelline} 
        & --\cellcolor{isabelline}
        & 43.9\cellcolor{isabelline} 
        & --\cellcolor{isabelline} 
        \\
\cdashline{1-9}
\rule{-4pt}{1.05\normalbaselineskip}
\multirow{2}{*}{\modelnameme{}}
    & \multicolumn{1}{c}{0} 
        & 76.3 
        & 61.0 
        & 64.2 
        & 47.8 
        & 28.8 
        & 15.6 
        & 22.6 
        \\
    & \multicolumn{1}{c}{8} 
        & \textbf{116.3}\cellcolor{isabelline} 
        & \textbf{106.6}\cellcolor{isabelline} 
        & \textbf{88.2 }\cellcolor{isabelline} 
        & \textbf{63.6}\cellcolor{isabelline} 
        & \textbf{46.3}\cellcolor{isabelline} 
        & \textbf{45.3}\cellcolor{isabelline} 
        & \textbf{51.4}\cellcolor{isabelline} 
        \\
\midrule[0.06em]
\multicolumn{9}{l}{\emph{MM1-30B Model Comparisons}} \\
\midrule[0.06em]
\multirow{3}{*}{IDEFICS-80B~\cite{obelics}}
    & \multicolumn{1}{c}{~0$^\dagger$}  
        & 91.8* 
        & 65.0 
        & 56.8 
        & 60.0 
        & 30.9 
        & 36.0 
        & 45.2 
        \\
    & \multicolumn{1}{c}{8} 
        & 114.3*\cellcolor{isabelline} 
        & 105.7\cellcolor{isabelline} 
        & 77.6\cellcolor{isabelline} 
        & 64.8\cellcolor{isabelline} 
        & 35.7\cellcolor{isabelline} 
        & 46.1\cellcolor{isabelline} 
        & 55.1\cellcolor{isabelline} 
        \\
    & \multicolumn{1}{c}{16} 
        & 116.6*\cellcolor{lightblue} 
        & 107.0\cellcolor{lightblue} 
        & 81.4\cellcolor{lightblue} 
        & 65.4\cellcolor{lightblue} 
        & 36.3\cellcolor{lightblue} 
        & 48.3\cellcolor{lightblue} 
        & 56.8\cellcolor{lightblue} 
        \\
\cdashline{1-9}
\rule{-4pt}{1.05\normalbaselineskip}
\multirow{3}{*}{Flamingo-80B~\cite{flamingo}}
    & \multicolumn{1}{c}{0$^\dagger$}  
        & 84.3 
        & --
        & --
        & 56.3 
        & 35.0 
        & 31.6 
        & 50.6 
        \\
    & \multicolumn{1}{c}{8} 
        & 108.8\cellcolor{isabelline} 
        & --\cellcolor{isabelline}
        & --\cellcolor{isabelline}
        & 65.6\cellcolor{isabelline} 
        & 37.3\cellcolor{isabelline} 
        & 44.8\cellcolor{isabelline} 
        & 57.5\cellcolor{isabelline} 
        \\
    & \multicolumn{1}{c}{16} 
        & 110.5\cellcolor{lightblue} 
        & --\cellcolor{lightblue}
        & --\cellcolor{lightblue}
        & 66.8\cellcolor{lightblue} 
        & 37.6\cellcolor{lightblue} 
        & 48.4\cellcolor{lightblue} 
        & 57.8\cellcolor{lightblue} 
        \\
\cdashline{1-9}
\rule{-4pt}{1.05\normalbaselineskip}
\multirow{3}{*}{Emu2-37B~\cite{sun2023generative}}
    & \multicolumn{1}{c}{0} 
        & -- 
        & --
        & --
        & 33.3 
        & 26.2 
        & 40.4 
        & 26.7 
        \\
    & \multicolumn{1}{c}{8} 
        & --\cellcolor{isabelline} 
        & --\cellcolor{isabelline}
        & --\cellcolor{isabelline}
        & 67.8\cellcolor{isabelline} 
        & 49.3\cellcolor{isabelline} 
        & 54.7\cellcolor{isabelline} 
        & 54.1\cellcolor{isabelline} 
        \\
    & \multicolumn{1}{c}{16} 
        &\cellcolor{lightblue} -- 
        &\cellcolor{lightblue} --
        &\cellcolor{lightblue} --
        & 68.8\cellcolor{lightblue} 
        & 50.3\cellcolor{lightblue} 
        & 57.0\cellcolor{lightblue} 
        & 57.1\cellcolor{lightblue} 
        \\
\cdashline{1-9}
\rule{-4pt}{1.05\normalbaselineskip}
\multirow{3}{*}{\modelnamelarge{}} 
    & \multicolumn{1}{c}{0} 
        & 70.3 
        & 54.6 
        & 64.9 
        & 48.9 
        & 28.2 
        & 14.5 
        & 24.1 
        \\
    & \multicolumn{1}{c}{8} 
        & 123.1\cellcolor{isabelline} 
        & 111.6\cellcolor{isabelline} 
        & 92.9\cellcolor{isabelline} 
        & 70.9\cellcolor{isabelline} 
        & 49.4\cellcolor{isabelline} 
        & 49.9\cellcolor{isabelline} 
        & 58.3\cellcolor{isabelline} 
        \\
    & \multicolumn{1}{c}{16} 
        & \textbf{125.3}\cellcolor{lightblue} 
        & \textbf{116.0}\cellcolor{lightblue} 
        & \textbf{97.6}\cellcolor{lightblue} 
        & \textbf{71.9}\cellcolor{lightblue} 
        & \textbf{50.6}\cellcolor{lightblue} 
        & \textbf{57.9}\cellcolor{lightblue} 
        & \textbf{59.3}\cellcolor{lightblue} 
        \\
\bottomrule
\end{tabular}}
\vspace{1mm}
\caption{Multimodal pre-training evaluations. (*) IDEFICS includes PMD in its training data (includes COCO). ($\dag$) These models include two text-only demonstrations in their ``0'' prompt, whereas \modelname{} does not. For the full table, see Table~\ref{tab:pt-results} in Appendix.}
\label{tab:pt-results-short}
\vspace*{-5mm}
\end{table*}

\vspace{1mm}
\noindent\textbf{Multimodal Pre-training Results.}
We evaluate pre-trained models on captioning and VQA tasks via appropriate prompting.\footnote{The models are prompted with ``\texttt{\{IMAGE\} A photo of}'' for captioning, and ``\texttt{\{IMAGE\} Question: \{QUESTION\} Short answer:}'' for VQA. See Appendix~\ref{app:pretrain-eval} for more details on pre-training evaluation.} We evaluate zero- and few-shot, as shown in Table \ref{tab:pt-results-short}, and compare against the few approaches that report few-shot pre-training performance. Note that we only compare our model with larger models, \emph{e.g.}, comparing our 30B model with two 80B models. 

When it comes to few-shot performance, \modelname{} outperforms all published prior work for pre-trained MLLMs. We see superior performance at 30B across captioning benchmarks and the VizWiz-QA benchmark. On VQAv2, TextVQA, OKVQA, at that scale we are comparable to Emu2~\cite{sun2023generative}.  
For zero-shot performance\footnote{We provide zero-shot results as a reference for the associated few-shot numbers, but we intentionally do not hill-climb on zero-shot metrics as they are mostly indicative of how well the pre-training mixture matches the associated evaluation task format.}, even without instruction fine-tuning, our models perform favorably on TextCaps across all model sizes, and comparable to Flamingo-3B at small scales for most benchmarks. 
\section{Supervised Fine-Tuning}
\label{sec:sft}

In this section, we describe the supervised fine-tuning (SFT) experiments trained on top of the pre-trained models described in the previous sections.
 
\vspace{1mm}
\noindent\textbf{SFT Data Mixture.} We follow LLaVA-1.5~\cite{liu2023improved} and LLaVA-NeXT~\cite{liu2024llavanext}, and collect roughly 1.45M SFT examples from a diverse set of datasets, including
\begin{itemize}[noitemsep, leftmargin=*]
    \item Instruction-response pairs generated by GPT-4 and GPT-4V, including LLaVA-Conv and LLaVA-Complex~\cite{llava} for conversations and complex reasoning, and ShareGPT-4V~\cite{chen2023sharegpt4v}\footnote{We also experimented with LVIS-Instruct4V~\cite{wang2023see}, but did not observe better performance than using ShareGPT-4V~\cite{chen2023sharegpt4v}, thus it is not included in the final mixture.} for detailed image descriptions;
    \item Academic task oriented vision-language (VL) datasets, including ($i$) VQAv2~\cite{goyal2017making}, GQA~\cite{hudson2019gqa},  OKVQA~\cite{marino2019ok}, A-OKVQA~\cite{schwenk2022okvqa}, and COCO Captions~\cite{chen2015microsoft} for natural images; ($ii$) OCRVQA~\cite{mishra2019ocr}, and TextCaps~\cite{sidorov2020textcaps} for text-rich images; and ($iii$) DVQA~\cite{kafle2018dvqa}, ChartQA~\cite{masry2022chartqa}, AI2D~\cite{kembhavi2016diagram}, DocVQA~\cite{mathew2021docvqa}, InfoVQA~\cite{mathew2022infographicvqa}, and Synthdog-En~\cite{kim2022ocr} for document and chart understanding.
    \item Text-only SFT data: We include an internal text-only dataset to ensure the model is capable of text-only instruction following.
\end{itemize}
The academic VL datasets are formatted into the instruction-following format, following LLaVA-1.5~\cite{liu2023improved}. More details are provided in Appendix \ref{app:sft_datasets}. All datasets are mixed together and randomly sampled during training.\footnote{While some different data mixing strategies were explored, simply mixing these datasets already achieves good performance, similar to observations in Honeybee~\cite{cha2023honeybee}.}

During SFT, we keep both the image encoder and the LLM backbone \emph{unfrozen}; other SFT training details are provided in Appendix~\ref{app:sft_training_details}. We evaluate our models across 12 benchmarks (see Appendix~\ref{app:benchmarks} for details). 

\vspace{1mm}
\noindent\textbf{Scaling to Higher Resolutions.} 
Intuitively, higher image resolution leads to better performance. 
To support high-resolution SFT, we use two approaches: 

    \textbf{Positional embedding interpolation}, \emph{e.g.}, as explored in Qwen-VL~\cite{bai2023qwen} and BLIP2~\cite{li2023blip}. After positional embedding interpolation, the vision transformer backbone is adapted to the new resolution during fine-tuning. Through this method, we have fine-tuned our model to support image resolutions ranging from $448\times448$, $560\times560$, to $672\times672$. Note that, for a resolution of $672\times672$, with a patch size of $14\times14$, an image is represented with $2,304$ tokens.
    
    \textbf{Sub-image decomposition}, recently introduced by SPHINX~\cite{lin2023sphinx}, Monkey~\cite{li2023monkey}, and LLaVA-NeXT~\cite{liu2024llavanext}. Computing self-attention among more than $2,000$ image tokens is computationally challenging, limiting further scaling to even higher image resolutions. Following SPHINX~\cite{lin2023sphinx}, as shown in Figure~\ref{fig:sphinx}, for a high-resolution input image, \emph{e.g.}, $1344\times1344$, we construct five images of $672\times672$, and feed them as independent images into our visual encoder. Specifically, we first downsample the input image to $672\times672$ as a high-level representation, and also resize the input image to $1344\times1344$ and divide the resized image into 4 sub-images of $672\times672$, which preserve more detailed visual information. Using positional embedding interpolation for each sub-image, we can support image resolution as high as $1792\times1792$ in experiments.

\begin{table*}[t]
\centering
\resizebox{\linewidth}{!}{  
\begin{tabular}{l | ccc | cccccc ccc}
\toprule
Model      & VQA$^\text{v2}$ & VQA$^\text{T}$ & SQA$^\text{I}$                                        & MMMU & MathV     & MME$^\text{P}$ & MME$^\text{C}$ & MMB   & SEED        & POPE  & LLaVA$^\text{W}$ & MM-Vet \\
\midrule
\multicolumn{11}{l}{\emph{3B Model Comparison}} \\
\midrule
MobileVLM~\cite{chu2023mobilevlm}      & --              & 47.5         & 61.0              & --/--  & --     & 1288.9         & --             & 59.6  & --/--       & 84.9  & --               & --               \\
LLaVA-Phi~\cite{zhu2024llava}        & 71.4            & 48.6              & 68.4                 & --/--   & --    & 1335.1         & --             & 59.8  & --/--       & 85.0  & --               & 28.9            \\
Imp-v1~\cite{imp2024}    & 79.45           & 59.38             & 69.96                           & --/--  & --      & 1434.0         & --             & 66.49 & --          & 88.02 & --               & 33.1            \\
TinyLLaVA~\cite{zhou2024tinyllava}    & 79.9            & 59.1              & 69.1               & --/--  & --    & 1464.9         & --             & 66.9  & --/--       & 86.4  &  75.8             & 32.0              \\
Bunny~\cite{he2024efficient}    & 79.8            & --                & 70.9                   & 38.2/33.0 & --  & 1488.8         & 289.3          & 68.6  & 62.5/--     & 86.8  & --               & --                 \\
Gemini Nano-2~\cite{team2023gemini}    & 67.5            & 65.9              & --                 & 32.6/-- & 30.6  & --             & --             & --    & --          & --    & --               & --              \\
\rowcolor{lightblue}
\modelnamesmallchat{}       &  82.0           & 71.9              & 69.4                                      & 33.9/33.7 & 32.0  & 1482.5         & 279.3          & 67.8  & 63.0/68.8    & 87.4  & 72.1             & 43.7          \\
\rowcolor{isabelline}
MM1-3B-MoE-Chat   &   82.5         & 72.9             & 76.1          & 38.6/35.7   & 32.6 &  1469.4         & 303.1         & 70.8  & 63.9/69.4    & 87.6  &  76.8           & 42.2            \\
\midrule
\multicolumn{11}{l}{\emph{7B Model Comparison}} \\
\midrule
InstructBLIP-7B~\cite{instruct-blip}    & --              & 50.1         & 60.5              & --/--  & 25.3    & --             & --             & 36.0  & 53.4/--     & --    & 60.9             & 26.2           \\
Qwen-VL-Chat-7B~\cite{bai2023qwen}    & 78.2            & 61.5              & 68.2              & 35.9/32.9 & --  & 1487.5         & 360.7          & 60.6  & 58.2/65.4   & --    & --               & --                 \\
LLaVA-1.5-7B~\cite{liu2023improved}   & 78.5            & 58.2              & 66.8               & --/--   & --    & 1510.7         & 316.1          & 64.3  & 58.6/66.1   & 85.9  & 63.4             & 31.1             \\
ShareGPT4V-7B~\cite{chen2023sharegpt4v}  & 80.6            & 60.4              & 68.4             & --/--  & --    & 1567.4         & 376.4          & 68.8  & --/--       & --    & 72.6             & --               \\
LVIS-Ins4V-7B~\cite{wang2023see}      & 79.6            & 58.7              & 68.3                & --/--   & --     & 1528.2         & --             & 66.2  & 60.6/--     & 86.0  & 67.0             & 31.5           \\
VILA-7B~\cite{lin2023vila}      & 79.9            & 64.4              & 68.2                          & --/--   & --     & 1531.3         & --             & 68.9  & 61.1/--     & 85.5  & 69.7             & 34.9       \\
SPHINX-Intern2~\cite{gao2024sphinx}  & 75.5            & --                & 70.4                 & --/-- & 35.5    & 1260.4         & 294.6          & 57.9  & 68.8/--     &  86.9  & 57.6             & 36.5            \\
LLaVA-NeXT-7B~\cite{liu2024llavanext}   & 81.8            & 64.9              & 70.1              & 35.8/-- & 34.6   & 1519           & 332            & 67.4  & --/70.2     & 86.53 & 81.6             & 43.9           \\
\rowcolor{lightblue} 
\modelnamechat{}         & 82.8           & 72.8             & 72.6                                & 37.0/35.6 &  35.9 & 1529.3         & 328.9          &  72.3  & 64.0/69.9 & 86.6  & 81.5             & 42.1              \\
\rowcolor{isabelline}
MM1-7B-MoE-Chat   &  83.4       & 73.8             & 74.4      & 40.9/37.9   & 40.9 &  1597.4         & 394.6         & 72.7  & 65.5/70.9    & 87.8  &   84.7         & 45.2           \\
\midrule
\multicolumn{11}{l}{\emph{30B Model Comparison}} \\
\midrule
Emu2-Chat-37B~\cite{sun2023generative}    & 84.9            & 66.6              & --             & 36.3/34.1  & -- & --             & --             & --    & 62.8/--     & --    & --               & 48.5             \\
CogVLM-30B~\cite{wang2023cogvlm}    & 83.4            & 68.1              & --                  & 32.1/30.1  & --   & --             & --             & --    & --          & --    & --               & 56.8             \\
LLaVA-NeXT-34B~\cite{liu2024llavanext}   & 83.7            & 69.5              & 81.8           & 51.1/44.7  & 46.5 & 1631           & 397            & 79.3  & --/75.9     & 87.73 & 89.6             & 57.4             \\
\rowcolor{lightblue} 
\modelnamelargechat{}  &  83.7     & 73.5               &  81.0         & 44.7/40.3  & 39.4$^\dagger$ &   1637.6         &  431.4          & 75.1  & 65.9/72.1    & 87.6 & 89.3            &     48.7         \\
\midrule
\rowcolor{magnolia}
Gemini Pro~\cite{team2023gemini}  & 71.2            & 74.6              & --                   & 47.9/-- & 45.2   & --             & 436.79         & 73.6  & --/70.7     & --    & --               & 64.3            \\
\rowcolor{magnolia}
Gemini Ultra~\cite{team2023gemini}    & 77.8            & 82.3              & --                      & 59.4/-- & 53.0  & --             & --             & --    & --          & --    & --               & --          \\
\rowcolor{magnolia}
GPT4V~\cite{achiam2023gpt}     & 77.2            & 78.0              & --                       & 56.8/55.7 & 49.9 & --             & 517.14         & 75.8  & 67.3/69.1   & --    & --               & 67.6              \\
\bottomrule
\end{tabular}
}
\vspace{1mm}
\caption{ \small
Comparison with SOTA models on MLLM benchmarks. VQA$^\text{v2}$~\cite{goyal2017making}; VQA$^\text{T}$: TextVQA~\cite{singh2019towards}; SQA$^\text{I}$: ScienceQA-IMG~\cite{lu2022learn}; MMMU~\cite{yue2023mmmu}; MathV: MathVista~\cite{lu2023mathvista}; MME$^\text{P/C}$: the Perception/Cognition split of MME~\cite{fu2023mme}; MMB: MMBench~\cite{liu2023mmbench}; SEED: SEED-Bench~\cite{li2023seed}; POPE~\cite{li2023evaluating}; LLaVA$^\text{W}$: LLaVA-Bench (In-the-Wild)~\cite{llava}; MM-Vet~\cite{yu2023mm}. The two numbers reported in MMMU denote the performance on the val and test split, respectively. The two numbers reported in SEED denote the performance on the whole SEED-Bench and the image part, respectively. ($\dagger$) 8-shot prompting: 44.4.
}
\label{tab:sota_comparison}
\vspace{-6mm}
\end{table*}

\vspace{-2mm}
\subsection{SFT Results} \label{subsec:sft_ablations}

\noindent\textbf{Comparison with SOTA.}
Results are summarized in Table \ref{tab:sota_comparison}. We use ``-Chat'' to denote our MM1 models after SFT. First, on average, MM1-3B-Chat and MM1-7B-Chat outperforms all listed models of the same size, setting a new state of the art for these model sizes.
MM1-3B-Chat and MM1-7B-Chat show particularly strong performance on VQAv2, TextVQA, ScienceQA, and also the more recent benchmarks (MMMU and MathVista). 

Second, we explore two MoE models: ($i$) 3B-MoE with 64 experts, and ($ii$) 7B-MoE with 32 experts. Our MoE models achieve uniformly better performance than the dense counterpart on almost every benchmark. This shows the great potential of MoE for further scaling, which is left as future work.

Third, for the 30B model size, MM1-30B-Chat outperforms Emu2-Chat-37B~\cite{sun2023generative} and CogVLM-30B~\cite{wang2023cogvlm} on TextVQA, SEED, and MMMU. Compared with the concurrent LLaVA-NeXT~\cite{liu2024llavanext}, we also achieve competitive performance across the board.
However, LLaVA-NeXT does not support multi-image reasoning, nor few-shot prompting, as each image is represented as 2,880 tokens sent to the LLM, while ours is only 720 in total. This limits certain applications that involve multiple images.

\begin{figure}[tb]
\centering
\begin{subfigure}[b]{0.32\textwidth}
    \includegraphics[width=\textwidth]{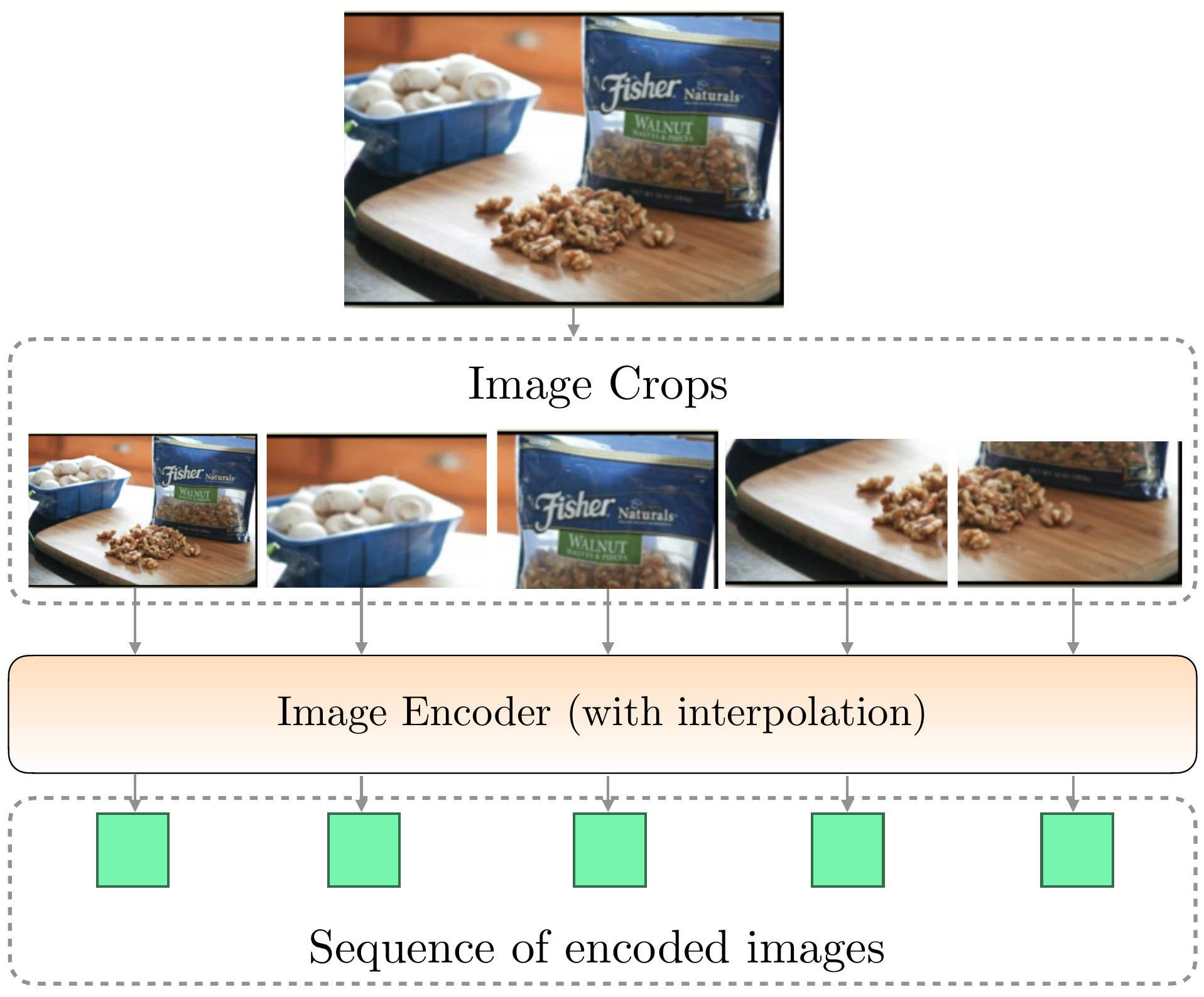}
    \vspace{0.36cm}
    \caption{High resolution image input processing.}
    \label{fig:sphinx}
\end{subfigure}\hfill
  \begin{subfigure}[b]{0.32\textwidth}
         \centering
         {
\pgfplotsset{compat=newest}
\begin{tikzpicture}
  \begin{axis}[
    xmode=log,
    xlabel=Image resolution,
    ylabel style={align=center}, ylabel=Average evaluation metric,
    grid=both,
    grid style={line width=.1pt, draw=gray!10},
    major grid style={line width=.2pt,draw=gray!50},
    minor tick num=5,
    axis line style={latex-latex},
    xmin=200,
    xmax=2000,
    ymin=75,
    ymax=100,
    width=1.15\textwidth,
    height=1.15\textwidth,
    xtick={224,336,448,672,896,1344,1792},
    xticklabels={224,336,448,672,896,1344,1792},
    xticklabel style={font=\tiny},
    tick label style={font=\scriptsize},
        yticklabel style = {font=\tiny},
        xticklabel style = {font=\tiny},
        ylabel style = {font=\tiny},
        xlabel style = {font=\tiny},
    xticklabel style={rotate=45},
    xlabel shift=-3pt
    ]
    \addplot +[line width=1.5pt,color=orange,mark options={fill=orange}] coordinates {
        (224,	78.50)
        (336,	87.14)
        (448,	89.40)
        (560,	92.02)
        (672,	97.97)
        (896,	99.75)
        (1344,	100)
        (1792,	98.54)
    };
  \end{axis}
\end{tikzpicture}\\
}
         \caption{Impact of image resolution on SFT performance.}
         \label{fig:sft-resolution}
     \end{subfigure}
     \hfill
     \begin{subfigure}[b]{0.32\textwidth}
         \centering
         \pgfplotsset{compat=newest}
\begin{tikzpicture}
  \begin{axis}[
    xlabel=Pre-training step (1000s),
    ylabel style={align=center}, ylabel=Average evaluation metric,
    grid=both,
    grid style={line width=.1pt, draw=gray!10},
    major grid style={line width=.2pt,draw=gray!50},
    minor tick num=5,
    axis line style={latex-latex},
    xmin=0,
    xmax=200,
    ymin=92,
    ymax=100,
    width=1.15\textwidth,
    height=1.15\textwidth,
        yticklabel style = {font=\tiny},
        xticklabel style = {font=\tiny},
        ylabel style = {font=\tiny,yshift=-1.5ex},
        xlabel style = {font=\tiny},    
    ]
    \addplot +[line width=1.5pt,color=orange,mark options={fill=orange}] coordinates {
        (5,	93.1)
        (45,	95.9)
        (105,	98.6)
        (145,	98.7)
        (200,	100)
    };
  \end{axis}
\end{tikzpicture}\\[2ex]
         \caption{Impact of pre-training on SFT performance.}
         \label{fig:sft_pretraining_step}
     \end{subfigure}
    \caption{We study the impact of image resolution and pre-training for SFT performance.}
    \label{fig:resolution_pretrain}
\vspace{-2mm}
\end{figure}

\vspace{1mm}
\noindent\textbf{Impact of Image Resolution.}
Figure~\ref{fig:sft-resolution} shows the impact of input image resolution on the average performance of the SFT evaluation metrics (defer the details of how we calculate the meta-average to Appendix~\ref{app:sft-meta-average}). Compared to a baseline model with an image resolution of 336 pixels, we can achieve a 15\% relative increase by supporting an image resolution of $1344\times1344$. Note that for the largest image resolution of $1792\times1792$, average performance decreases slightly. This is likely because many of the evaluation images are smaller than this resolution, and resizing artifacts may affect the model performance. By default, the results in Table \ref{tab:sota_comparison} correspond to image resolutions of $1344\times1344$.

\vspace{1mm}
\noindent\textbf{Impact of Pre-training.}
In contrast to most recent MLLMs, we perform large-scale pre-training for our models. To assess the impact of pre-training on the final model performance, we perform SFT on the same pre-training run, but at different checkpoint steps. For an earlier checkpoint step, the model has seen less unique data samples than a later checkpoint step, so this is a measure of the importance of the quantity of pre-training data.  In Figure~\ref{fig:sft_pretraining_step}, we show that the model consistently improves as it has seen more pre-training data. Furthermore, large-scale multimodal pre-training enables strong in-context few-shot learning and multi-image reasoning capabilities, while most MLLM benchmarks shown in Table \ref{tab:sota_comparison} focus on zero-shot metrics and single-image reasoning. 

\vspace{1mm}
\noindent\textbf{Few-shot Chain-of-Thought Reasoning after SFT.}
As seen in Section~\ref{sec:data}, \modelname{} gains few-shot capabilities thanks to interleaved data. Even though our fine-tuning data includes only single-image examples, we find that \modelnamelargechat{} still exhibits multi-image reasoning. This is shown qualitatively in Figure \ref{fig:qualitative_example_mm1_sft}, and quantitatively on MathVista~\cite{lu2023mathvista}, where we evaluate few-shot performance with chain-of-thought prompting: 4-shot performance is \textbf{41.9}, which is 2.5 points higher than zero-shot (\textbf{39.4}).

Our best performing high-resolution SFT model uses 720 tokens per image. This is a challenge when using more than 4 in-context examples due to the context length. To allow for more examples, we explore a \emph{mixed resolution in-context examples} formulation, where we feed some of the examples at a lower resolution (see Appendix~\ref{app:fewshot_sft} for details). Using this formulation with 8 in-context examples increases the performance on MathVista to \textbf{44.4}.

\vspace{1mm}
\noindent\textbf{Do the lessons learned via pre-training transfer to SFT?} Yes. We find that (1) pre-training with caption-only data improves SFT metrics, and (2) different VL connector architectures have negligible impact on final results.  Detailed ablation results are provided in Appendix~\ref{app:additional_sft_ablation}. 

\vspace{1mm}
\noindent\textbf{Qualitative Analysis.} To better understand MM1, more qualitative examples are provided in Appendix \ref{app:qualitative}, including single-image and multi-image reasoning, and few-shot prompting.
\vspace{-2mm}
\section{Conclusion}
\vspace{-2mm}
We study how to build performant \MLMabbr{}s. Through carefully ablating modeling and data choices, we identify important lessons that yield a pre-trained model achieving SOTA results on a range of few-shot evaluations. After SFT, this model family produces competitive performance on a wide range of benchmarks, while enabling multi-image reasoning and few-shot prompting. We hope that the identified lessons will help the community in building strong models beyond any single specific model architecture or data strategy.

\bibliographystyle{splncs04}
\bibliography{egbib}
\clearpage
\appendix
\startcontents[sections]
\setcounter{tocdepth}{2}
\renewcommand*\contentsname{Appendix}

\section*{Appendix}
\printcontents[sections]{l}{1}{\setcounter{tocdepth}{2}}

\section{Dataset Details} \label{app: dataset_details}

\subsection{Interleaved Image-Text Data}\label{sec:app:interleaved}

Following a process similar to OBELICS~\cite{obelics}, we construct a dataset of 500M interleaved image-text documents, containing 1B images and 500B text tokens. These 500M documents are built from a collection of 3B HTML files described in Sec. \ref{sec:app:textonly}. From each of the HTML files, we extract the text body layer and all the \texttt{<img>} tags. We remove documents that have no images or more than 30 images. We then download the images and insert them at their original positions in the text. Finally, we perform \textbf{image filtering} and \textbf{image de-duplication} to remove low-quality and repetitive images.

During image filtering, we remove images that have corrupted bytes and/or header, aspect ratio less than 1/2 or greater than 2, are too small (less than 100px) or too large (larger than 10,000px), or if their URL contains \textit{logo}, \textit{button}, \textit{icon}, \textit{plugin} or \textit{widget}.
During image de-duplication, we remove images whose URL or MD5 hash have appeared more than 10 times in the dataset. Additionally, when an image appears multiple times on a single page, we only retain its first appearance.

\subsection{Text-Only Data}\label{sec:app:textonly}

From an initial Web corpus of 150B English HTML files, we perform boilerplate removal to arrive at the HTML representing the main content. We then follow similar processes as GPT-3~\cite{brown2020language} and CCNet~\cite{wenzek2019ccnet} to filter out documents that are too short, contain profanity, or are otherwise considered low-quality documents. We de-duplicate the data using exact-hash matching and LSH-based near-duplicate detection.  Using these methods, we arrive at 3B HTML files.

\subsection{Visual Instruction Tuning Data}\label{app:sft_datasets}

\begin{table}[t]
\small
    \centering
    \begin{tabular}{l  | l | l  }
        \toprule
        \bf{Datasets} & \bf{Size} & \bf{Prompting Strategy} \\
        \midrule
        Text-only SFT & 13k &  -- \\
        \midrule
        LLaVA-Conv~\cite{llava} & 57k &   \\
        LLaVA-Complex~\cite{llava} & 77k &  -- \\
        ShareGPT-4V~\cite{chen2023sharegpt4v}  & 102k & \\
        \midrule
        VQAv2~\cite{goyal2017making}   & 83k  &  \multirow{8}{*}{\makecell[l]{``Answer the question using  a single word or \quad \quad \\ phrase.''}} \\
        GQA~\cite{hudson2019gqa}     & 72k & \\
        OKVQA~\cite{marino2019ok}   & 9k  & \\
        OCRVQA~\cite{mishra2019ocr}  & 80k & \\
        DVQA~\cite{kafle2018dvqa}     & 200k & \\
        ChartQA~\cite{masry2022chartqa}	& 18k	& \\
        AI2D~\cite{kembhavi2016diagram}    & 3k  & \\
        DocVQA~\cite{mathew2021docvqa}  & 39k	& \\
        InfoVQA~\cite{mathew2022infographicvqa} & 24k	& \\ 
        \midrule
        A-OKVQA~\cite{schwenk2022okvqa} & 66k & \makecell[l]{``Answer with the option's letter from the given \\ choices directly.''} \\
        \midrule
        COCO Captions~\cite{chen2015microsoft} & 83k  & \multirow{2}{*}{\makecell[l]{Sample from a pre-generated prompt list, \emph{e.g.}, \quad \quad \\ ``Provide a brief description of the given image.''}}\\
        TextCaps~\cite{sidorov2020textcaps} & 22k  & \\
        \midrule
        SynthDog-EN~\cite{kim2022ocr} & 500k & \makecell[l]{Sample from a pre-generated prompt list, \emph{e.g.}, \\ ``Please transcribe all the text in the picture.''}\\ 
        \midrule
        \textbf{Total}  & \textbf{1.45M }& -- \\
        \bottomrule
    \end{tabular}
    \vspace{1mm}
    \caption{List of datasets used for supervised fine-tuning.}
    \label{tab:sft_data_mixture_statistics}
    \vspace{-5mm}
\end{table}

Our final SFT data mixture contains a variety of datasets, mostly follow LLaVA-1.5~\cite{liu2023improved} and LLaVA-NeXT~\cite{liu2024llavanext}. Specifically, 
\begin{itemize}[noitemsep, leftmargin=*]
    \item To encourage the model to provide long-form detailed responses and perform conversations, we follow previous work, use the existing GPT-4 generated data (LLaVA-Conv and LLaVA-Complex~\cite{llava}) and the existing GPT-4V generated data (ShareGPT-4V~\cite{chen2023sharegpt4v}) for model training. We also experimented with LAION-GPT4V, but did not observe further performance improvement, thus not included in the final mixture.
    \item To enhance the model with better vision-language (VL) understanding capability, we use a variety of academic task oriented VL datasets. These datasets are either in the form of image captioning, or in the form of VQA with short answers. Specifically,
    \begin{itemize}[noitemsep, leftmargin=*]
        \item For natural images: VQAv2~\cite{goyal2017making}, GQA~\cite{hudson2019gqa},  OKVQA~\cite{marino2019ok}, A-OKVQA~\cite{schwenk2022okvqa}, and COCO Captions~\cite{chen2015microsoft};
        \item For text-rich images: OCRVQA~\cite{mishra2019ocr}, and TextCaps~\cite{sidorov2020textcaps};
        \item For document and chart understanding: DVQA~\cite{kafle2018dvqa}, ChartQA~\cite{masry2022chartqa}, AI2D~\cite{kembhavi2016diagram}, DocVQA~\cite{mathew2021docvqa}, InfoVQA~\cite{mathew2022infographicvqa}, and SynthDog-En~\cite{kim2022ocr};
    \end{itemize}
    \item To enhance the model's text-only instruction following capability, we also blend in a small amount of text-only SFT data.
\end{itemize}
The academic task oriented image captioning and VQA datasets are formatted into the instruction-following format, following LLaVA-1.5~\cite{liu2023improved}, with detailed prompts summarized in Table \ref{tab:sft_data_mixture_statistics}.

\section{Training Details}\label{app:training-details}

\subsection{Pre-training}\label{app:pre-training}





\begin{table}[htbp]
\centering
\small
\resizebox{0.95\linewidth}{!}{  
\begin{tabular}{lcccccccc}
\toprule 
\multirow{2}{*}{\bf Model} 
    & \multirow{2}{*}{~\bf Shot~} 
    & \multicolumn{3}{c}{\bf Captioning} 
    & \multicolumn{4}{c}{\bf Visual Question Answering} 
    \\
\cmidrule(lr){3-5} \cmidrule(lr){6-9}
&
    & COCO 
    & NoCaps 
    & TextCaps
    & VQAv2 
    & TextVQA 
    & VizWiz 
    & OKVQA \\ \midrule
\multicolumn{9}{l}{\emph{MM1-3B Model Comparisons}} \\
\midrule
\multirow{4}{*}{Flamingo-3B~\cite{flamingo}}
    & ~0$^\dag$
        & 73.0\cellcolor{zeroshot!50} 
        & --\cellcolor{zeroshot!50}
        & --\cellcolor{zeroshot!50}
        & 49.2\cellcolor{zeroshot!50} 
        & 30.1\cellcolor{zeroshot!50} 
        & 28.9\cellcolor{zeroshot!50} 
        & 41.2\cellcolor{zeroshot!50} 
        \\
    & 4
        & 85.0\cellcolor{fourshot!50} 
        & --\cellcolor{fourshot!50}
        & --\cellcolor{fourshot!50}
        & 53.2\cellcolor{fourshot!50} 
        & 32.7\cellcolor{fourshot!50} 
        & 34.0\cellcolor{fourshot!50} 
        & 43.3\cellcolor{fourshot!50} 
        \\
    & 8
        & 90.6\cellcolor{eightshot!50} 
        & --\cellcolor{eightshot!50}
        & --\cellcolor{eightshot!50}
        & 55.4\cellcolor{eightshot!50} 
        & 32.4\cellcolor{eightshot!50} 
        & 38.4\cellcolor{eightshot!50} 
        & 44.6\cellcolor{eightshot!50} 
        \\
    & 16
        & 95.3\cellcolor{sixteenshot!50} 
        & --\cellcolor{sixteenshot!50}
        & --\cellcolor{sixteenshot!50}
        & 56.7\cellcolor{sixteenshot!50} 
        & 31.8\cellcolor{sixteenshot!50} 
        & 43.3\cellcolor{sixteenshot!50} 
        & 45.6\cellcolor{sixteenshot!50} 
        \\ \cdashline{1-9}
\rule{-4pt}{1.05\normalbaselineskip}
\multirow{4}{*}{\modelnamesmall{}}
    & 0
        & 73.5\cellcolor{zeroshot!50} 
        & 55.6\cellcolor{zeroshot!50} 
        & 63.3\cellcolor{zeroshot!50} 
        & 46.2\cellcolor{zeroshot!50} 
        & 29.4\cellcolor{zeroshot!50} 
        & 15.6\cellcolor{zeroshot!50} 
        & 26.1\cellcolor{zeroshot!50} 
        \\
  & 4
        & 112.3\cellcolor{fourshot!50} 
        & 99.7\cellcolor{fourshot!50} 
        & 84.1\cellcolor{fourshot!50} 
        & 57.9\cellcolor{fourshot!50} 
        & 45.3\cellcolor{fourshot!50} 
        & 38.0\cellcolor{fourshot!50} 
        & 48.6\cellcolor{fourshot!50} 
        \\
 & 8
        & 114.6\cellcolor{eightshot!50} 
        & 104.7\cellcolor{eightshot!50} 
        & 88.8\cellcolor{eightshot!50} 
        & 63.6\cellcolor{eightshot!50} 
        & 44.6\cellcolor{eightshot!50} 
        & 46.4\cellcolor{eightshot!50} 
        & 48.4\cellcolor{eightshot!50} 
        \\
 & 16
        & 116.8\cellcolor{sixteenshot!50} 
        & 107.6\cellcolor{sixteenshot!50} 
        & 91.6\cellcolor{sixteenshot!50} 
        & 60.9\cellcolor{sixteenshot!50} 
        & 46.1\cellcolor{sixteenshot!50} 
        & 53.8\cellcolor{sixteenshot!50} 
        & 50.5\cellcolor{sixteenshot!50} 
        \\
\midrule
\multicolumn{9}{l}{\emph{MM1-7B Model Comparisons}} \\
\midrule
\multirow{4}{*}{IDEFICS-9B~\cite{obelics}}
    & ~0$^\dag$
        & 46.0*\cellcolor{zeroshot!50} 
        & 36.8\cellcolor{zeroshot!50} 
        & 25.4\cellcolor{zeroshot!50} 
        & 50.9\cellcolor{zeroshot!50} 
        & 25.9\cellcolor{zeroshot!50} 
        & 35.5\cellcolor{zeroshot!50} 
        & 38.4\cellcolor{zeroshot!50} 
        \\
    & 4
        & 93.0*\cellcolor{fourshot!50} 
        & 81.3\cellcolor{fourshot!50} 
        & 60.0\cellcolor{fourshot!50} 
        & 55.4\cellcolor{fourshot!50} 
        & 27.6\cellcolor{fourshot!50} 
        & 36.9\cellcolor{fourshot!50} 
        & 45.4\cellcolor{fourshot!50} 
        \\
    & 8
        & 97.0*\cellcolor{eightshot!50} 
        & 86.8\cellcolor{eightshot!50}
        & 63.2\cellcolor{eightshot!50} 
        & 56.4\cellcolor{eightshot!50} 
        & 27.5\cellcolor{eightshot!50} 
        & 40.4\cellcolor{eightshot!50} 
        & 47.7\cellcolor{eightshot!50} 
        \\
    & 16
        & 99.7*\cellcolor{sixteenshot!50} 
        & 89.4\cellcolor{sixteenshot!50} 
        & 67.4\cellcolor{sixteenshot!50} 
        & 57.0\cellcolor{sixteenshot!50} 
        & 27.9\cellcolor{sixteenshot!50} 
        & 42.6\cellcolor{sixteenshot!50} 
        & 48.4\cellcolor{sixteenshot!50} 
        \\ \cdashline{1-9}
\rule{-4pt}{1.05\normalbaselineskip}
\multirow{4}{*}{Flamingo-9B~\cite{flamingo}}
    & ~0$^\dag$
        & 79.4\cellcolor{zeroshot!50} 
        & --\cellcolor{zeroshot!50}
        & --\cellcolor{zeroshot!50}
        & 51.8\cellcolor{zeroshot!50} 
        & 31.8\cellcolor{zeroshot!50} 
        & 28.8\cellcolor{zeroshot!50} 
        & 44.7\cellcolor{zeroshot!50} 
        \\
    & 4
        & 93.1 \cellcolor{fourshot!50} 
        & --\cellcolor{fourshot!50}
        & --\cellcolor{fourshot!50}
        & 56.3\cellcolor{fourshot!50} 
        & 33.6\cellcolor{fourshot!50} 
        & 34.9\cellcolor{fourshot!50} 
        & 49.3\cellcolor{fourshot!50} 
        \\
    & 8
        & 99.0\cellcolor{eightshot!50} 
        & --\cellcolor{eightshot!50}
        & --\cellcolor{eightshot!50}
        & 58.0\cellcolor{eightshot!50} 
        & 33.6\cellcolor{eightshot!50} 
        & 39.4\cellcolor{eightshot!50} 
        & 50.0\cellcolor{eightshot!50} 
        \\
    & 16
        & 102.2\cellcolor{sixteenshot!50} 
        & --\cellcolor{sixteenshot!50}
        & --\cellcolor{sixteenshot!50}
        & 59.4\cellcolor{sixteenshot!50}  
        & 33.5\cellcolor{sixteenshot!50} 
        & 43.0\cellcolor{sixteenshot!50} 
        & 50.8\cellcolor{sixteenshot!50} 
        \\ \cdashline{1-9}
\rule{-4pt}{1.05\normalbaselineskip}
\multirow{3}{*}{Emu2-14B~\cite{sun2023generative}}
    & ~0$^\dag$
        &\cellcolor{zeroshot!50} -- 
        & --\cellcolor{zeroshot!50}
        & --\cellcolor{zeroshot!50}
        & 52.9\cellcolor{zeroshot!50} 
        & --\cellcolor{zeroshot!50}
        & 34.4\cellcolor{zeroshot!50} 
        & 42.8\cellcolor{zeroshot!50} 
        \\
    & 4
        &\cellcolor{fourshot!50} -- 
        &\cellcolor{fourshot!50} --
        &\cellcolor{fourshot!50} --
        & 58.4\cellcolor{fourshot!50} 
        & --\cellcolor{fourshot!50}
        & 41.3\cellcolor{fourshot!50} 
        & -- \cellcolor{fourshot!50} 
        \\
    & 8
        &\cellcolor{eightshot!50} -- 
        & --\cellcolor{eightshot!50}
        & --\cellcolor{eightshot!50}
        & 59.0\cellcolor{eightshot!50} 
        & --\cellcolor{eightshot!50}
        & 43.9 \cellcolor{eightshot!50} 
        & -- \cellcolor{eightshot!50} 
        \\ \cdashline{1-9}
\rule{-4pt}{1.05\normalbaselineskip}
\multirow{4}{*}{\modelnameme{}}
    & 0
        & 76.3\cellcolor{zeroshot!50} 
        & 61.0\cellcolor{zeroshot!50} 
        & 64.2\cellcolor{zeroshot!50} 
        & 47.8\cellcolor{zeroshot!50} 
        & 28.8\cellcolor{zeroshot!50} 
        & 15.6\cellcolor{zeroshot!50} 
        & 22.6\cellcolor{zeroshot!50} 
        \\
    & 4
        & 109.8\cellcolor{fourshot!50} 
        & 96.2\cellcolor{fourshot!50} 
        & 84.5\cellcolor{fourshot!50} 
        & 60.6\cellcolor{fourshot!50} 
        & 44.4\cellcolor{fourshot!50} 
        & 37.4\cellcolor{fourshot!50} 
        & 46.6\cellcolor{fourshot!50} 
        \\
    & 8
        & 116.3\cellcolor{eightshot!50} 
        & 106.6\cellcolor{eightshot!50} 
        & 88.2\cellcolor{eightshot!50} 
        & 63.6\cellcolor{eightshot!50} 
        & 46.3\cellcolor{eightshot!50} 
        & 45.3\cellcolor{eightshot!50} 
        & 51.4\cellcolor{eightshot!50} 
        \\
    & 16
        & 118.6\cellcolor{sixteenshot!50} 
        & 111.1\cellcolor{sixteenshot!50} 
        & 93.1\cellcolor{sixteenshot!50} 
        & 65.2\cellcolor{sixteenshot!50} 
        & 46.9\cellcolor{sixteenshot!50} 
        & 53.2\cellcolor{sixteenshot!50} 
        & 52.9\cellcolor{sixteenshot!50} 
        \\
\midrule[0.06em]
\multicolumn{9}{l}{\emph{MM1-30B Model Comparisons}} \\
\midrule[0.06em]
\multirow{4}{*}{IDEFICS-80B~\cite{obelics}}
    & ~0$^\dag$
        & 91.8*\cellcolor{zeroshot!50} 
        & 65.0 \cellcolor{zeroshot!50}
        & 56.8\cellcolor{zeroshot!50} 
        & 60.0\cellcolor{zeroshot!50} 
        & 30.9\cellcolor{zeroshot!50} 
        & 36.0\cellcolor{zeroshot!50} 
        & 45.2 \cellcolor{zeroshot!50} 
        \\
    & 4
        & 110.3*\cellcolor{fourshot!50} 
        & 99.6\cellcolor{fourshot!50} 
        & 72.7\cellcolor{fourshot!50} 
        & 63.6\cellcolor{fourshot!50} 
        & 34.4\cellcolor{fourshot!50} 
        & 40.4\cellcolor{fourshot!50} 
        & 52.4\cellcolor{fourshot!50} 
        \\
    & 8
        & 114.3*\cellcolor{eightshot!50} 
        & 105.7\cellcolor{eightshot!50} 
        & 77.6\cellcolor{eightshot!50} 
        & 64.8\cellcolor{eightshot!50} 
        & 35.7\cellcolor{eightshot!50} 
        & 46.1\cellcolor{eightshot!50} 
        & 55.1\cellcolor{eightshot!50} 
        \\
    & 16
        & 116.6*\cellcolor{sixteenshot!50} 
        & 107.0\cellcolor{sixteenshot!50} 
        & 81.4\cellcolor{sixteenshot!50} 
        & 65.4\cellcolor{sixteenshot!50} 
        & 36.3\cellcolor{sixteenshot!50} 
        & 48.3\cellcolor{sixteenshot!50} 
        & 56.8\cellcolor{sixteenshot!50} 
        \\ \cdashline{1-9}
\rule{-4pt}{1.05\normalbaselineskip}
\multirow{4}{*}{Flamingo-80B~\cite{flamingo}}
    & ~0$^\dag$
        & 84.3\cellcolor{zeroshot!50} 
        & --\cellcolor{zeroshot!50}
        & --\cellcolor{zeroshot!50}
        & 56.3\cellcolor{zeroshot!50} 
        & 35.0\cellcolor{zeroshot!50} 
        & 31.6\cellcolor{zeroshot!50} 
        & 50.6\cellcolor{zeroshot!50} 
        \\
    & 4
        & 103.2\cellcolor{fourshot!50} 
        & --\cellcolor{fourshot!50}
        & --\cellcolor{fourshot!50}
        & 63.1\cellcolor{fourshot!50} 
        & 36.5\cellcolor{fourshot!50} 
        & 39.6\cellcolor{fourshot!50} 
        & 57.4\cellcolor{fourshot!50} 
        \\
    & 8
        & 108.8\cellcolor{eightshot!50} 
        & --\cellcolor{eightshot!50}
        & --\cellcolor{eightshot!50}
        & 65.6\cellcolor{eightshot!50} 
        & 37.3\cellcolor{eightshot!50} 
        & 44.8\cellcolor{eightshot!50} 
        & 57.5\cellcolor{eightshot!50} 
        \\
    & 16
        & 110.5\cellcolor{sixteenshot!50} 
        & --\cellcolor{sixteenshot!50}
        & --\cellcolor{sixteenshot!50}
        & 66.8\cellcolor{sixteenshot!50} 
        & 37.6\cellcolor{sixteenshot!50} 
        & 48.4\cellcolor{sixteenshot!50} 
        & 57.8\cellcolor{sixteenshot!50} 
        \\ \cdashline{1-9}
\rule{-4pt}{1.05\normalbaselineskip}
\multirow{4}{*}{Emu2-37B~\cite{sun2023generative}}
    & 0
        &\cellcolor{zeroshot!50} -- 
        & --\cellcolor{zeroshot!50}
        & --\cellcolor{zeroshot!50}
        & 33.3\cellcolor{zeroshot!50} 
        & 26.2\cellcolor{zeroshot!50} 
        & 40.4\cellcolor{zeroshot!50} 
        & 26.7\cellcolor{zeroshot!50} 
        \\
    & 4
        &\cellcolor{fourshot!50} -- 
        & --\cellcolor{fourshot!50}
        & --\cellcolor{fourshot!50}
        & 67.0\cellcolor{fourshot!50} 
        & 48.2\cellcolor{fourshot!50} 
        & 54.6\cellcolor{fourshot!50} 
        & 53.2\cellcolor{fourshot!50} 
        \\
    & 8
        &\cellcolor{eightshot!50} -- 
        & --\cellcolor{eightshot!50}
        & --\cellcolor{eightshot!50}
        & 67.8\cellcolor{eightshot!50} 
        & 49.3\cellcolor{eightshot!50} 
        & 54.7\cellcolor{eightshot!50} 
        & 54.1\cellcolor{eightshot!50} 
        \\
    & 16
        &\cellcolor{sixteenshot!50} -- 
        & --\cellcolor{sixteenshot!50}
        & --\cellcolor{sixteenshot!50}
        & 68.8\cellcolor{sixteenshot!50} 
        & 50.3\cellcolor{sixteenshot!50} 
        & 57.0\cellcolor{sixteenshot!50} 
        & 57.1\cellcolor{sixteenshot!50} 
        \\ \cdashline{1-9}
\rule{-4pt}{1.05\normalbaselineskip}
\multirow{4}{*}{\modelnamelarge{}} 
    & 0 
        & 70.3\cellcolor{zeroshot!50} 
        & 54.6\cellcolor{zeroshot!50} 
        & 64.9\cellcolor{zeroshot!50} 
        & 48.9\cellcolor{zeroshot!50} 
        & 28.2\cellcolor{zeroshot!50} 
        & 14.5\cellcolor{zeroshot!50} 
        & 24.1\cellcolor{zeroshot!50} 
        \\
    & 4
        & 117.9\cellcolor{fourshot!50} 
        & 103.8\cellcolor{fourshot!50} 
        & 87.5\cellcolor{fourshot!50} 
        & 68.8\cellcolor{fourshot!50} 
        & 48.1\cellcolor{fourshot!50} 
        & 41.7\cellcolor{fourshot!50} 
        & 54.9\cellcolor{fourshot!50} 
        \\
    & 8
        & 123.1\cellcolor{eightshot!50} 
        & 111.6\cellcolor{eightshot!50} 
        & 92.9\cellcolor{eightshot!50} 
        & 70.9\cellcolor{eightshot!50} 
        & 49.4\cellcolor{eightshot!50} 
        & 49.9\cellcolor{eightshot!50} 
        & 58.3\cellcolor{eightshot!50} 
        \\
    & 16
        & 125.3\cellcolor{sixteenshot!50} 
        & 116.0\cellcolor{sixteenshot!50} 
        & 97.6\cellcolor{sixteenshot!50} 
        & 71.9\cellcolor{sixteenshot!50} 
        & 50.6\cellcolor{sixteenshot!50} 
        & 57.9\cellcolor{sixteenshot!50} 
        & 59.3\cellcolor{sixteenshot!50} 
        \\
\bottomrule
\end{tabular}}
\vspace{1mm}
\caption{Complete \modelname{} pre-training few-shot evaluation results. (*) IDEFICS includes PMD in its training data (includes COCO). (\dag) These models included two text-only demonstrations in their ``0'' prompt, whereas \modelname{} does not.}
\label{tab:pt-results}
\vspace*{-7mm}
\end{table}

\noindent \textbf{Batch Size and Composition.}
For simplicity, all \modelname{} models are pre-trained with the same batch size of 512 and maximum decoder sequence length of 4096. We allow up to 16 images per input sequence, with each image resulting in 144 tokens as input to the decoder. Note that this results in roughly 1M text tokens and 1M image tokens per batch. Each input sequence is sampled from one of three types of input sources: (1) interleaved, (2) packed image-text pairs, or (3) text-only data, with sampling probability 45\%, 45\%, and 10\%, respectively. When packing image-text pairs or interleaved documents along the sequence dimension, we modify the self-attention masks to prevent tokens from attention across example boundaries. For image-text pairs in particular, this was critical for maintaining strong few-shot performance. 

\begin{wraptable}[12]{r}{0.4\textwidth}
    \vspace{-7mm}
    \centering
    \begin{tabular}{ccc}
    \toprule
        N & Pred. $\eta$ & Pred. $\lambda$ \\
    \midrule
        1.2B & 8.6e-5 & 5.0e-6 \\
        2.9B & 5.9e-5 & 3.5e-6 \\ 
        6.4B & 4.2e-5 & 2.5e-6 \\ 
        30B & 2.2e-5 & 1.3e-6 \\
    \bottomrule
    \end{tabular}
    \caption{Predicted optimal peak learning rate $\eta$ and weight decay $\lambda$ for MM1 model sizes.}
    \label{tab:pred-peak-lrs}
\end{wraptable}

Note that our sampling/mixing procedure is performed once offline and stored as a fixed \emph{deterministic} snapshot of our pre-training mixture. This means, with the exception of our ablations on the pre-training mixture itself, all models in this paper are trained on the same examples in the same order. We found this was critical to ensure internal reproducibility of our results, as initial experiments showed that different random seeds in the input pipeline could have non-negligible impact on resulting models.

\begin{wrapfigure}{r}{0.4\textwidth}
    \vspace{-11mm}
    \centering
    \include{figures/pretraining_scaling/wd_opt_fit}
    \vspace{-4mm}
    \caption{Optimal weight decay as a function of model size for the grid searches described in Sec.~\ref{app:pre-training}. The x-axis is the number of (non-embedding) LLM parameters and the y-axis is weight decay.}\label{fig:wd-opt-fit}
    \vspace{-10mm}
\end{wrapfigure}

\vspace{1mm}
\noindent \textbf{Learning Rate Schedule.} 
For multimodal pre-training, \modelname{} employs a standard cosine learning rate decay schedule with an initial linear warmup of 2000 steps. The learning rate is then decayed to 10\% of its peak value over the course of $2e5$ training steps. We perform gradient clipping with max norm 1 and use the AdamW optimizer with an implementation that decouples the learning rate and weight decay. For \modelnamelarge{}, we also add a z-loss term with scale 1e-4, as we observed this improves training stability, similar to \cite{smallscaleproxies}. 

The predicted optimal (peak) learning rates for each of the main LLM sizes studied in this work are shown in Table~\ref{tab:pred-peak-lrs}. For simplicity, for the actual \modelname{} 3B, 7B, and 30B models, we used $\eta$ equal to 6e-5, 4e-5, and 2e-5, respectively. Finally, we fix the peak LR of the randomly initialized vision-language connector of \modelname{} to $\eta=$8e-5 for all model sizes. For future versions of \modelname{}, we plan on incorporating techniques similar to \cite{mup} to avoid the need to conduct costly hyperparameter searches.

\begin{figure}[t]
    \centering
    \begin{subfigure}{0.2\textwidth}
        \centering
        \include{figures/pretraining_scaling/appendix/lr_wd_grid_9m}
         \caption{9M}
         \label{fig:lr-wd-grid-9m}
     \end{subfigure}
     \hspace*{3cm}
     \begin{subfigure}{0.2\textwidth}
         \centering
         \include{figures/pretraining_scaling/appendix/lr_wd_grid_85m}
         \caption{85M}
         \label{fig:lr-wd-grid-85m}
     \end{subfigure}
    \\
    \begin{subfigure}{0.2\textwidth}
         \centering
         \include{figures/pretraining_scaling/appendix/lr_wd_grid_302m}
         \caption{302M}
         \label{fig:lr-wd-grid-302m}
     \end{subfigure}
     \hspace*{3cm}
    \begin{subfigure}{0.2\textwidth}
         \centering
         \include{figures/pretraining_scaling/appendix/lr_wd_grid_1_2b}
         \caption{1.2B}
         \label{fig:lr-wd-grid-1-2b}
     \end{subfigure}
    \caption{8-shot average for grid searches over peak learning rate (y-axis) and weight decay (x-axis) for different LLM sizes. Black cells correspond to settings we did not run a corresponding experiment for.}
    \label{fig:lr-wd-grid-search}
    \vspace{-3mm}
\end{figure}

\noindent \textbf{Learning Rate and Weight Decay Grid Searches.}
The individual grid search results corresponding to the final curve fit in Figure~\ref{fig:lr-opt-fit} are shown in Figure~\ref{fig:lr-wd-grid-search}. We train grid search models for $5e^4$ steps, as \cite{smallscaleproxies} found this does not alter the conclusions. We can apply the same procedure that was used for predicting optimal learning rate to predict weight decay values, as shown in Figure~\ref{fig:wd-opt-fit}. The blue circles correspond to actual data points from the grid search with sampling probability (and darkness of color) proportional to their 8-shot average performance. The corresponding predictions for each of the main model sizes in this work are shown in Table~\ref{tab:pred-peak-lrs}.

\vspace{-2mm}
\subsection{Supervised Fine-tuning (SFT)}
\vspace{-1mm}
\label{app:sft_training_details}
The model is fine-tuned for 10k steps with batch size 256 and sequence length 2048. We employ the AdaFactor optimizer with peak learning rate 1e-5 and cosine decay to 0. We experimented different learning rates; empirically, the value of 1e-5 is optimal. During SFT, we keep both the image encoder and the LLM \emph{unfrozen}, as empirically, we observe that finetuning the whole model achieves better performance. 

\section{Evaluation Details}

\subsection{Pre-training Evaluation}\label{app:pretrain-eval}

\begin{wraptable}{r}{0.35\textwidth}
    \vspace{-15mm}
    \centering
    \small 
    \begin{tabular}{cc}
        \toprule
        Dataset &  Evaluation Split \\ \midrule
        COCO & Karpathy test \\
        NoCaps & val \\ 
        TextCaps & val \\ 
        VQAv2 & testdev \\ 
        TextVQA & val \\ 
        VizWiz & testdev \\ 
        OKVQA & val  \\ \bottomrule
    \end{tabular}
    \caption{Splits used for pre-training evaluation. Note that, unlike the main pre-training results, all pre-training ablations use the validation splits for VQAv2 and VizWiz.}
    \label{tab:pretraining_eval_splits}
\end{wraptable}

Few-shot prompts are randomly sampled per-dataset from the training set if available, otherwise the validation set (ensuring the query example does not appear in any of the shots). Outputs are generated with greedy decoding until the model emits the EOS token or any additional stop tokens that can be specified on a per-task basis. The additional stop token for captioning tasks is just the newline character, and for VQA tasks we also include ``.'', ``,'', and ``Question'' as valid stop tokens. For postprocessing VQA predictions, we use the same logic as OpenFlamingo\footnote{Specifcally, the implementation of \href{https://github.com/mlfoundations/open_flamingo/blob/60a5fd6a6bf0940ccf0eba1c777d55b7306ccc53/open_flamingo/eval/vqa\_metric.py\#L210}{VQAMetric} (commit 60a5fd6).}\cite{open-flamingo}. For captioning tasks, we report CIDEr score~\cite{cider} using the nlg-eval package~\cite{sharma2017nlgeval}. All of our multimodal pre-training evaluations are implemented in an internal fork of EleutherAI's lm-evaluation-harness~\cite{eval-harness}.

\subsection{SFT Evaluation Benchmarks} \label{app:benchmarks}

We evaluate our SFT models on a collection of both traditional academic VL benchmarks and recent benchmarks specifically designed for MLLMs. For academic VL benchmarks, we include VQAv2~\cite{goyal2017making}, 
TextVQA~\cite{singh2019towards}, and the image subset of ScienceQA~\cite{lu2022learn}. For recent MLLM benchmarks, we include POPE~\cite{li2023evaluating}, MME~\cite{fu2023mme}, MMBench~\cite{liu2023mmbench}, SEED-Bench~\cite{li2023seed}, LLaVA-Bench-in-the-Wild~\cite{llava}, MM-Vet~\cite{yu2023mm}, MathVista~\cite{lu2023mathvista}, and the recent popular MMMU~\cite{yue2023mmmu}. For all the benchmarks, we use greedy decoding to generate the responses. For MM-Vet and LLaVA-Bench-in-the-Wild, which use GPT-4 for evaluation, we run the evaluation 3 times, and report the average. 

\subsection{SFT Evaluation Meta-Average}\label{app:sft-meta-average}

In the process of SFT ablation, we synthesize all benchmark results into a single meta-average number to simplify comparisons. Because the evaluation metrics of different datasets may have different ranges, we normalize with respect to a baseline configuration. This is achieved by initially standardizing the results for each task; that is, we adjust every metric by dividing it by its respective baseline, followed by averaging across all metrics. To elaborate, we establish our baseline using the performance metrics of a compact MM1 model, which is trained on $224\times224$ image resolution and employs attention pooling with 64 image queries.

\subsection{Additional SFT Ablations}\label{app:additional_sft_ablation}
In this section, we perform SFT ablations. This section is analogous to Section~\ref{sec:methodology_overview}; here, we perform SFT on the same checkpoints and evaluate if similar lessons hold true on SFT evaluations, instead of pre-training evaluations. Furthermore, we also study whether to keep the image encoder frozen or not during SFT. For all of these ablations, we train \modelnamesmallchat{}.

\vspace{1mm}
\noindent\textbf{Pre-training data mixture ablations.} In Figure~\ref{fig:sft_ablations:mix}, we compare the SFT performance with different weights for pre-training data. We see a similar trend when comparing with Figure~\ref{fig:data-ablations} for 0-shot evaluations. Pre-training with caption-only data gives the best performance across the SFT evaluation metrics. This corroborates \textbf{Data lesson 1}: caption data still lifts zero-shot performance for SFT evaluations. However, the SFT metrics do not measure few-shot performance, so the impact of the interleaved data is not noticeable in this table.

\begin{figure}[tb]
     \centering
     \begin{subfigure}[b]{0.32\textwidth}
         \centering
         \definecolor{bblue}{HTML}{4F81BD}
\definecolor{rred}{HTML}{C0504D}
\definecolor{ggreen}{HTML}{9BBB59}
\definecolor{ppurple}{HTML}{9F4C7C}

\definecolor{myorange}{RGB}{252,213,180}
\definecolor{myblue}{RGB}{184,204,228}
\definecolor{myred}{RGB}{230,184,183}
\definecolor{mypurple}{RGB}{204,192,218}
\definecolor{myturquoise}{RGB}{183,222,232}
\definecolor{mybrown}{RGB}{196,189,151}
\definecolor{mygray}{RGB}{217,217,217}
\definecolor{mygreen}{RGB}{216,228,188}
\definecolor{myblue2}{RGB}{141,180,226}

\definecolor{mpl_blue}{HTML}{769FC7}
\definecolor{mpl_orange}{HTML}{F3AA72}
\definecolor{mpl_green}{HTML}{85BB77}
\definecolor{mpl_red}{HTML}{D6756F}
\definecolor{mpl_purple}{HTML}{AF96CD}

\usetikzlibrary{patterns}
    \usetikzlibrary{
        matrix,
    }
    \pgfplotsset{
        compat=1.3,
    }

\makeatletter 
\makeatother 
\newdimen\LineSpace
\tikzset{
    line space/.code={\LineSpace=#1},
    line space=3pt
}

\begin{tikzpicture}
    \begin{axis}[
        width  = \textwidth*1.2,
        height = 4cm,
        major x tick style = transparent,
        ybar=0pt,
        bar width=9pt,
        ymajorgrids = true,
        ylabel = {Average Performance},
        symbolic x coords={SFT-Metrics},
        xtick = data,
        ylabel near ticks,
        xlabel near ticks,
        scaled y ticks = false,
        enlarge x limits=0.5,
        x tick label style={ align=center},
        ymin=95,
        ymax=105,
        legend cell align=left,
        legend style={at={(0.5,-0.2)},
            anchor=north,legend columns=3,
            font=\tiny,
            inner sep=1pt,
            },
        legend image code/.code={
        \draw [#1] (-0.07cm,-0.1cm) rectangle (0.15cm,0.08cm); },
        nodes near coords,
        every node near coord/.append style={font=\tiny, rotate=90, anchor=west},
        nodes near coords align={vertical},
        yticklabel style = {font=\tiny,xshift=0.5ex},
        xticklabel style = {font=\tiny,yshift=0.5ex},
        ylabel style = {font=\tiny,yshift=-1.5ex},
        title style = {font=\small\bfseries,yshift=-1.5ex,},
    ]

        \addplot[style={black,fill=mpl_blue,mark=none}]
            coordinates {(SFT-Metrics, 103.1)}; \label{plot:line1}

        \addplot[style={black,fill=mpl_orange,mark=none,postaction={pattern=my north east lines, line space=8pt}}]
            coordinates {(SFT-Metrics, 99.9)  };\label{plot:line2}
        
        \addplot[style={black,fill=mpl_green,mark=none,postaction={pattern=my north west lines, line space=8pt}}]
            coordinates {(SFT-Metrics, 99.5)  };
        \label{plot:line3}

        \addplot[style={black,fill=mpl_red,mark=none,postaction={pattern=vertical lines}}]
            coordinates {(SFT-Metrics, 99.5) };

        \addplot[style={black,fill=mpl_purple,mark=none,postaction={pattern=horizontal lines}}]
            coordinates {(SFT-Metrics, 97.4)  };

        \addplot[style={black,fill=mybrown,mark=none,,postaction={pattern=grid}}]
            coordinates {(SFT-Metrics, 97.3)  };

        \addplot[style={black,fill=mygreen,mark=none,postaction={pattern=crosshatch dots}}]
            coordinates {(SFT-Metrics, 99.6)  };

        \addplot[style={black,fill=myblue2,mark=none,postaction={pattern=crosshatch}}]
            coordinates {(SFT-Metrics, 100.0)  };

        \legend{0/100/0, 0/50/50, 33/33/33, 50/50/0, 50/0/50, 100/0/0, 42/42/14, 45/45/9}
    \end{axis}

\end{tikzpicture}\\[-6ex]
         \caption{Interleaved/Captions/Text data mixture.  }
         \label{fig:sft_ablations:mix}
     \end{subfigure}
     \hfill
     \begin{subfigure}[b]{0.32\textwidth}
         \centering
         \definecolor{bblue}{HTML}{4F81BD}
\definecolor{rred}{HTML}{C0504D}
\definecolor{ggreen}{HTML}{9BBB59}
\definecolor{ppurple}{HTML}{9F4C7C}

\definecolor{myorange}{RGB}{252,213,180}
\definecolor{myblue}{RGB}{184,204,228}
\definecolor{myred}{RGB}{230,184,183}
\definecolor{mypurple}{RGB}{204,192,218}
\definecolor{myturquoise}{RGB}{183,222,232}
\definecolor{mybrown}{RGB}{196,189,151}
\definecolor{mygray}{RGB}{217,217,217}
\definecolor{mygreen}{RGB}{216,228,188}
\definecolor{myblue2}{RGB}{141,180,226}

\definecolor{mpl_blue}{HTML}{769FC7}
\definecolor{mpl_orange}{HTML}{F3AA72}
\definecolor{mpl_green}{HTML}{85BB77}
\definecolor{mpl_red}{HTML}{D6756F}
\definecolor{mpl_purple}{HTML}{AF96CD}

\makeatletter 
\makeatother 
\newdimen\LineSpace
\tikzset{
    line space/.code={\LineSpace=#1},
    line space=3pt
}

    \usetikzlibrary{
        matrix,
    }
    \pgfplotsset{
        compat=1.3,
    }

\begin{tikzpicture}
    \begin{axis}[
        width  = \textwidth*1.15,
        height = 4cm,
        major x tick style = transparent,
        ybar=0pt,
        bar width=7pt,
        ymajorgrids = true,
        ylabel = {Average SFT-Metrics},
        symbolic x coords={224px 64tks,336px 64tks,336px 144tks},
        x tick label style  = {text width=1cm,align=center},
        xmin=224px 64tks, xmax=336px 144tks,
        xtick = data,
        ylabel near ticks,
        xlabel near ticks,
        scaled y ticks = false,
        enlarge x limits=0.25,
        ymin=95,
        ymax=115,
        legend cell align=left,
        legend style={at={(0.5,-0.25)},
            anchor=north,legend columns=2,
            font=\tiny,
            inner sep=1pt,
            },
        legend image code/.code={
        \draw [#1] (-0.07cm,-0.1cm) rectangle (0.15cm,0.08cm); },
        nodes near coords,
        every node near coord/.append style={font=\tiny, rotate=90, anchor=west},
        nodes near coords align={vertical},
        yticklabel style = {font=\tiny,xshift=0.5ex},
        xticklabel style = {font=\tiny,yshift=0.5ex},
        ylabel style = {font=\tiny,yshift=-1.5ex},
        title style = {font=\small\bfseries,yshift=-1.5ex,},
    ]

        \addplot[style={black,fill=mpl_blue,mark=none}]
            coordinates {(224px 64tks, 101.3) (336px 64tks, 105.1) (336px 144tks, 105.2)  }; 

        \addplot[style={black,fill=mpl_orange,mark=none,postaction={pattern=my north east lines, line space=8pt}}]
            coordinates {(224px 64tks, 100.0) (336px 64tks, 103.2) (336px 144tks, 106.2)  }; 
        
        \addplot[style={black,fill=mpl_green,mark=none,postaction={pattern=my north west lines, line space=8pt}}]
            coordinates {(224px 64tks, 100.4) (336px 64tks, 105.5) (336px 144tks, 107.8)  };

        \legend{Avg. Pool, Att. Pool, C-Abstractor}
        
    \end{axis}
\end{tikzpicture}\\[-3ex]
         \caption{Vision-language connector.}
         \label{fig:sft_ablations:bridge}
     \end{subfigure}
     \hfill
     \begin{subfigure}[b]{0.32\textwidth}
         \centering
          \definecolor{bblue}{HTML}{4F81BD}
\definecolor{rred}{HTML}{C0504D}
\definecolor{ggreen}{HTML}{9BBB59}
\definecolor{ppurple}{HTML}{9F4C7C}

\definecolor{myorange}{RGB}{252,213,180}
\definecolor{myblue}{RGB}{184,204,228}
\definecolor{myred}{RGB}{230,184,183}
\definecolor{mypurple}{RGB}{204,192,218}
\definecolor{myturquoise}{RGB}{183,222,232}
\definecolor{mybrown}{RGB}{196,189,151}
\definecolor{mygray}{RGB}{217,217,217}
\definecolor{mygreen}{RGB}{216,228,188}
\definecolor{myblue2}{RGB}{141,180,226}

\definecolor{mpl_blue}{HTML}{769FC7}
\definecolor{mpl_orange}{HTML}{F3AA72}
\definecolor{mpl_green}{HTML}{85BB77}
\definecolor{mpl_red}{HTML}{D6756F}
\definecolor{mpl_purple}{HTML}{AF96CD}

\makeatletter 
\makeatother 
\newdimen\LineSpace
\tikzset{
    line space/.code={\LineSpace=#1},
    line space=3pt
}

    \usetikzlibrary{
        matrix,
    }
    \pgfplotsset{
        compat=1.3,
    }

\begin{tikzpicture}
    \begin{axis}[
        width  = \textwidth*1.15,
        height = 4cm,
        major x tick style = transparent,
        ybar=0pt,
        bar width=7pt,
        ymajorgrids = true,
        ylabel = {Average SFT-Metrics},
        symbolic x coords={336px 144tks,672px 720tks,1344px 720tks},
        x tick label style  = {text width=1cm,align=center},
        xmin=336px 144tks, xmax=1344px 720tks,
        xtick = data,
        ylabel near ticks,
        xlabel near ticks,
        scaled y ticks = false,
        enlarge x limits=0.25,
        ymin=90,
        ymax=120,
        legend cell align=left,
        legend style={at={(0.5,-0.25)},
            anchor=north,legend columns=1,
            font=\tiny,
            inner sep=1pt,
            },
        legend image code/.code={
        \draw [#1] (-0.07cm,-0.1cm) rectangle (0.15cm,0.08cm); },
        nodes near coords,
        every node near coord/.append style={font=\tiny, rotate=90, anchor=west},
        nodes near coords align={vertical},
        yticklabel style = {font=\tiny,xshift=0.5ex},
        xticklabel style = {font=\tiny,yshift=0.5ex},
        ylabel style = {font=\tiny,yshift=-1.5ex},
        title style = {font=\tiny\bfseries,yshift=-1.5ex,},
    ]

        \addplot[style={black,fill=mpl_blue,mark=none}]
            coordinates {(336px 144tks, 100.0) (672px 720tks, 109.8) (1344px 720tks, 113.0)  }; 

        \addplot[style={black,fill=mpl_orange,mark=none,postaction={pattern=my north east lines, line space=8pt}}]
            coordinates {(336px 144tks, 102.2) (672px 720tks, 109.6) (1344px 720tks, 110.1)  }; 

        \legend{Unfrozen Encoder, Frozen Encoder}
        
    \end{axis}
\end{tikzpicture}\\[-3ex]
         \caption{(Un-)freezing image encoder during SFT.}
         \label{fig:sft_ablations:encoder}
     \end{subfigure}
    \caption{\textbf{SFT ablations.} (a) The impact of pre-training data mixture on SFT results. Here, $x/y/z$ means that $x\%$ of the data is interleaved, $y\%$ is captions, and $z\%$ is pure text. tks: the number of image tokens. (b) The impact of different vision-language connectors on SFT results. For both (a) and (b), we first pre-train \modelnamesmall{} with the ablated setting, and then perform SFT on the pre-trained models. (c) Freezing or unfreezing the image encoder during SFT. }
    \label{fig:sft_ablations}
    \vspace{-3mm}
\end{figure}

\vspace{1mm}
\noindent\textbf{Visual-language connector ablations.} In Figure~\ref{fig:sft_ablations:bridge}, we evaluate different visual-language connector configurations. This figure is similar to Figure~\ref{fig:visual_language_bridge_ablations}, except that we evaluate the corresponding SFT models. As can be seen, if a low number of image tokens is used, average pooling gives similar results as C-Abstractor. When the number of image tokens is increased, the C-Abstractor configuration gives the best results.
These trends are not entirely consistent with pre-training results reported in Figure~\ref{fig:visual_language_bridge_ablations}. Overall, the impact of the choice of visual-language connector appears to not have a very significant impact on final test performance. Our final models use the C-Abstractor architecture.

\vspace{1mm}
\noindent\textbf{Image encoder ablations.} 
In Figure~\ref{fig:sft_ablations:encoder}, we study whether to keep the image encoder frozen or not during SFT. The results show that at lower image resolutions, a frozen image encoder results in better performance than an unfrozen image encoder (+2.2 points). However, at higher resolutions (\emph{i.e.}, 1344px), it is beneficial to unfreeze the image encoder (+2.9 points). This is likely because the pre-training is performed at the base resolution without any interpolation or image sub-divisions.

\subsection{Implementation Details for Few-shot \modelnamelargechat{}}\label{app:fewshot_sft}
As shown in Section~\ref{subsec:sft_ablations}, our fine-tuned model can utilize in-context examples to achieve even stronger performance. Interestingly, the performance goes up when increasing the number of examples. We demonstrate this with \modelnamelargechat{}.

One challenge for few-shot inputs arises due to the use of sub-image decomposition. While this strategy lifts zero-shot performance, it significantly increases the effective number of tokens consumed per image. Using 5 sub-images per input image as \modelnamelargechat{} does, processing a 4-shot example where every example contains just one source image already yields 20 effective images. Representing every image with 144 tokens therefore requires 2,880 tokens for images alone, quickly exhausting limited language model context. To mitigate this limitation, we propose a new \emph{mixed-resolution} approach. Specifically, for $K$ in-context examples, we only encode the last $N$ images at a high resolution via sub-image decomposition, the remaining $K-N$ in-context examples are processed at lower resolution. This makes in-context examples much less expensive to encode and allows to increase the number of in-context examples within a fixed token budget, further increasing performance. In our implementation, we set $N=3$.

We demonstrate the effectiveness of our proposed strategy using the MathVista benchmark. Using in-context learning with chain-of-thought, the performance of \modelnamelargechat{} improves from 39.4 (0-shot) to 41.9 (4-shot, with all in-context examples using full sub-image decomposition). Applying our mixed-resolution approach allows to encode additional in-context examples, enabling up to 8-shot chain of thought, further improving the performance to 44.4.
This illustrates that our MM1-Chat model retains in-context learning capabilities inherited from its pre-training regimen, and that our strategy of mixed decomposition is effective at further increasing few-shot performance within a restricted context length.

\newpage
\section{Qualitative Examples}\label{app:qualitative}

In this section, we share qualitative examples of MM1 predictions.

\begin{figure}[h!]
\centering
{
\renewcommand{\arrayrulewidth}{0.4mm}
\renewcommand\arraystretch{1.5}
\setlength{\fboxsep}{0pt}
\setlength{\tabcolsep}{3pt}
\setlength{\fboxrule}{0.9pt}
\includegraphics[width=0.90\textwidth]{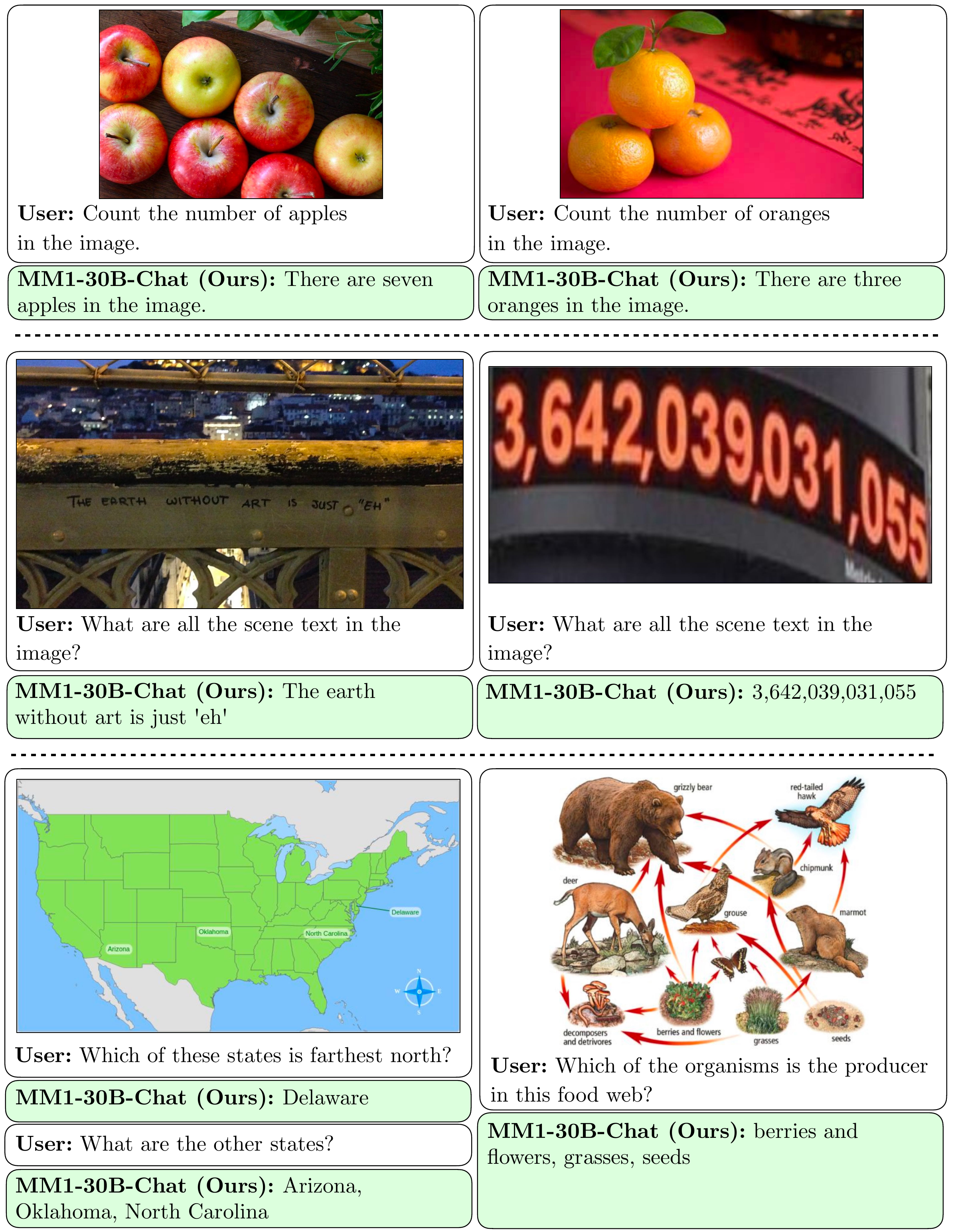}
}
\caption{Examples testing MM1 counting, OCR and scientific knowledge capabilities. Images and prompts are from COCO 2014 validation set \cite{cocodataset} and \cite{yang2023dawn}.}
    \label{fig:qualitative_example_appendix_example_3}
    \vspace{-5mm}
\end{figure}

\newpage

\begin{figure}[h!]
\centering
{
\renewcommand{\arrayrulewidth}{0.4mm}
\renewcommand\arraystretch{1.5}
\setlength{\fboxsep}{0pt}
\setlength{\tabcolsep}{3pt}
\setlength{\fboxrule}{0.9pt}
\includegraphics[width=0.90\textwidth]{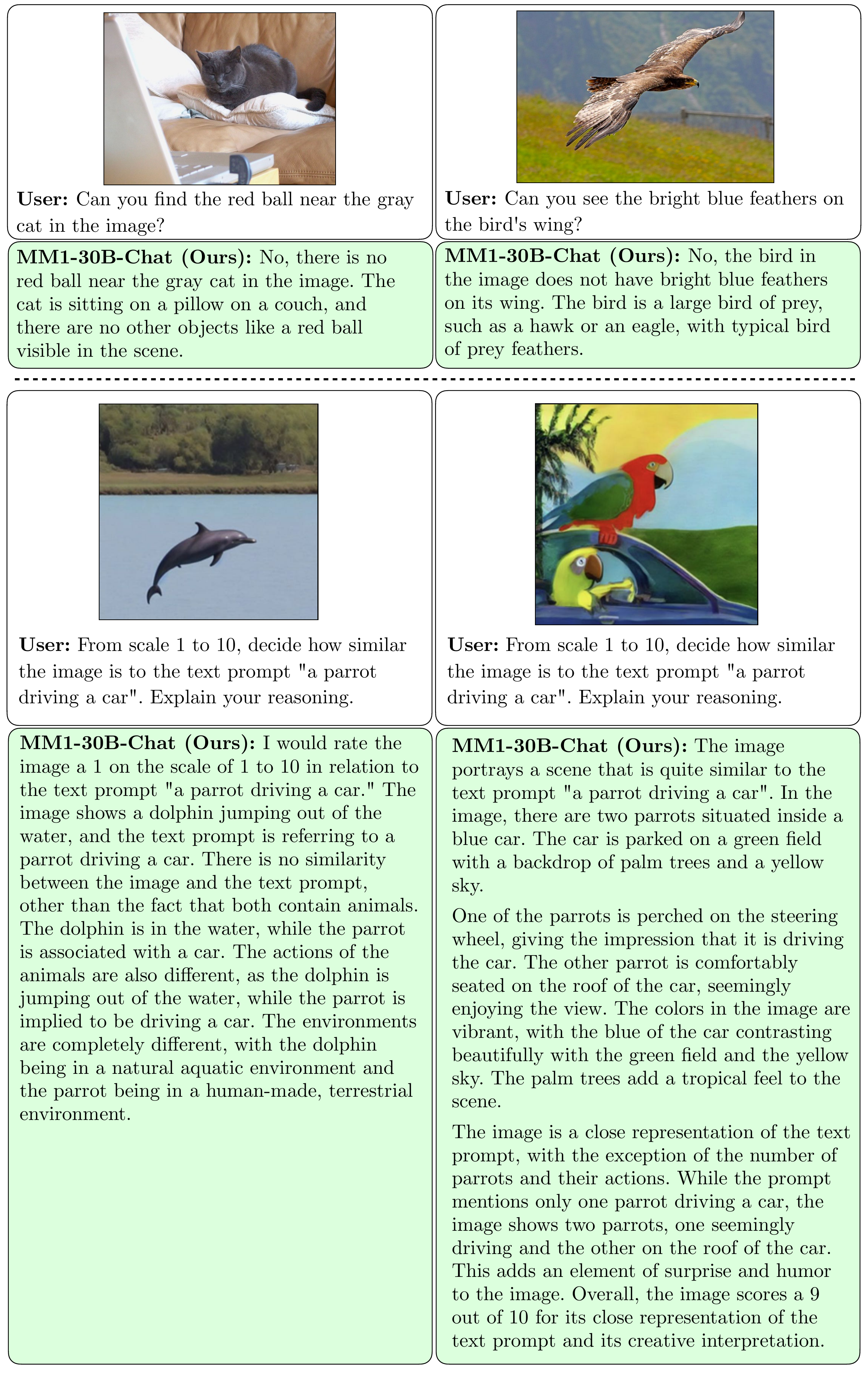}
}
\caption{Examples testing MM1 against adversarial prompts and image-prompt alignment. 
Images and prompts are from COCO 2014 validation set \cite{cocodataset}, \cite{yang2023dawn} and \cite{black2023training}.}
    \label{fig:qualitative_example_appendix_example_4}
    \vspace{-5mm}
\end{figure}

\newpage

\begin{figure}[h!]
\centering
{
\renewcommand{\arrayrulewidth}{0.4mm}
\renewcommand\arraystretch{1.5}
\setlength{\fboxsep}{0pt}
\setlength{\tabcolsep}{3pt}
\setlength{\fboxrule}{0.9pt}
\includegraphics[width=1\textwidth]{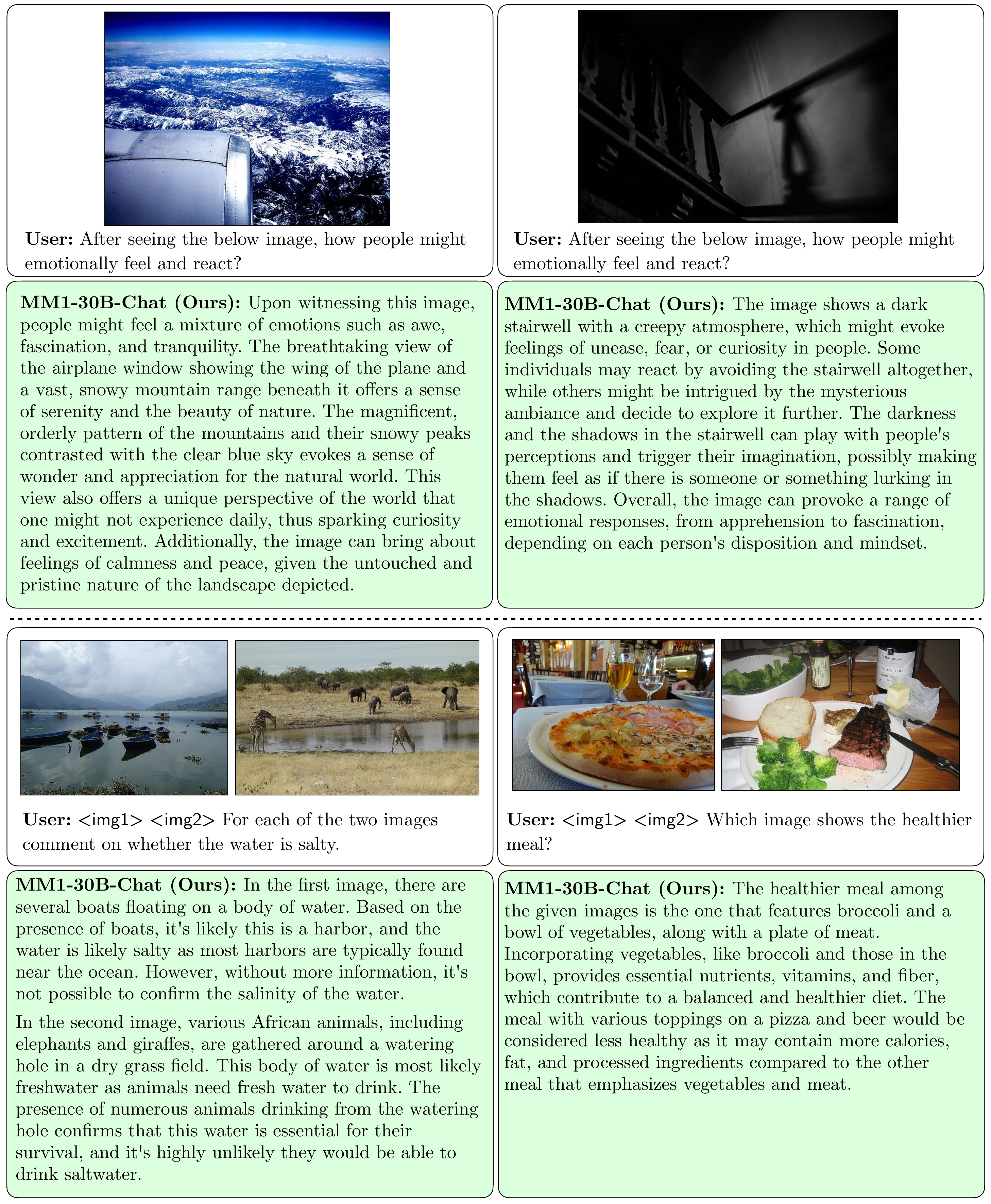}
}
\caption{Examples testing MM1 ability to perceive image aesthetics and compare multiple images. Images and prompts are from COCO 2014 validation set \cite{cocodataset} and \cite{yang2023dawn}.}
    \label{fig:qualitative_example_appendix_example_5}
    \vspace{-1.2cm}
\end{figure}

\begin{figure}[h!]
\centering
{
\renewcommand{\arrayrulewidth}{0.4mm}
\renewcommand\arraystretch{1.5}
\setlength{\fboxsep}{0pt}
\setlength{\tabcolsep}{3pt}
\setlength{\fboxrule}{0.9pt}
\includegraphics[width=1\textwidth]{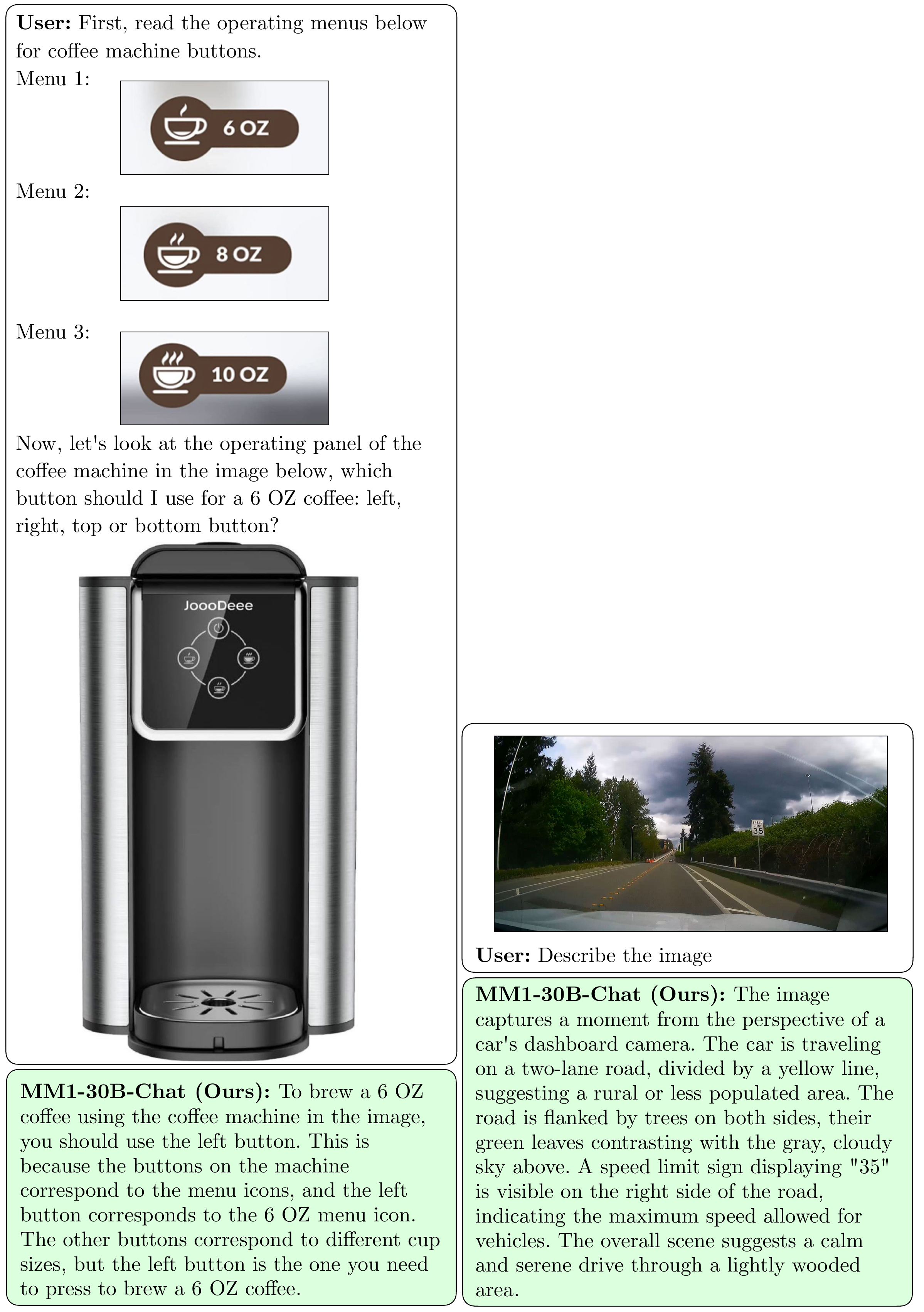}
}
\caption{Following \cite{yang2023dawn}, we tested MM1 on task-oriented scenarios such as operating machines and navigating. Images and prompts are from \cite{yang2023dawn}.}
    \label{fig:qualitative_example_appendix_example_6}
    \vspace{-5mm}
\end{figure}

\newpage

\begin{figure}[h!]
\centering
{
\renewcommand{\arrayrulewidth}{0.4mm}
\renewcommand\arraystretch{1.5}
\setlength{\fboxsep}{0pt}
\setlength{\tabcolsep}{3pt}
\setlength{\fboxrule}{0.9pt}
\includegraphics[width=1.00\textwidth]{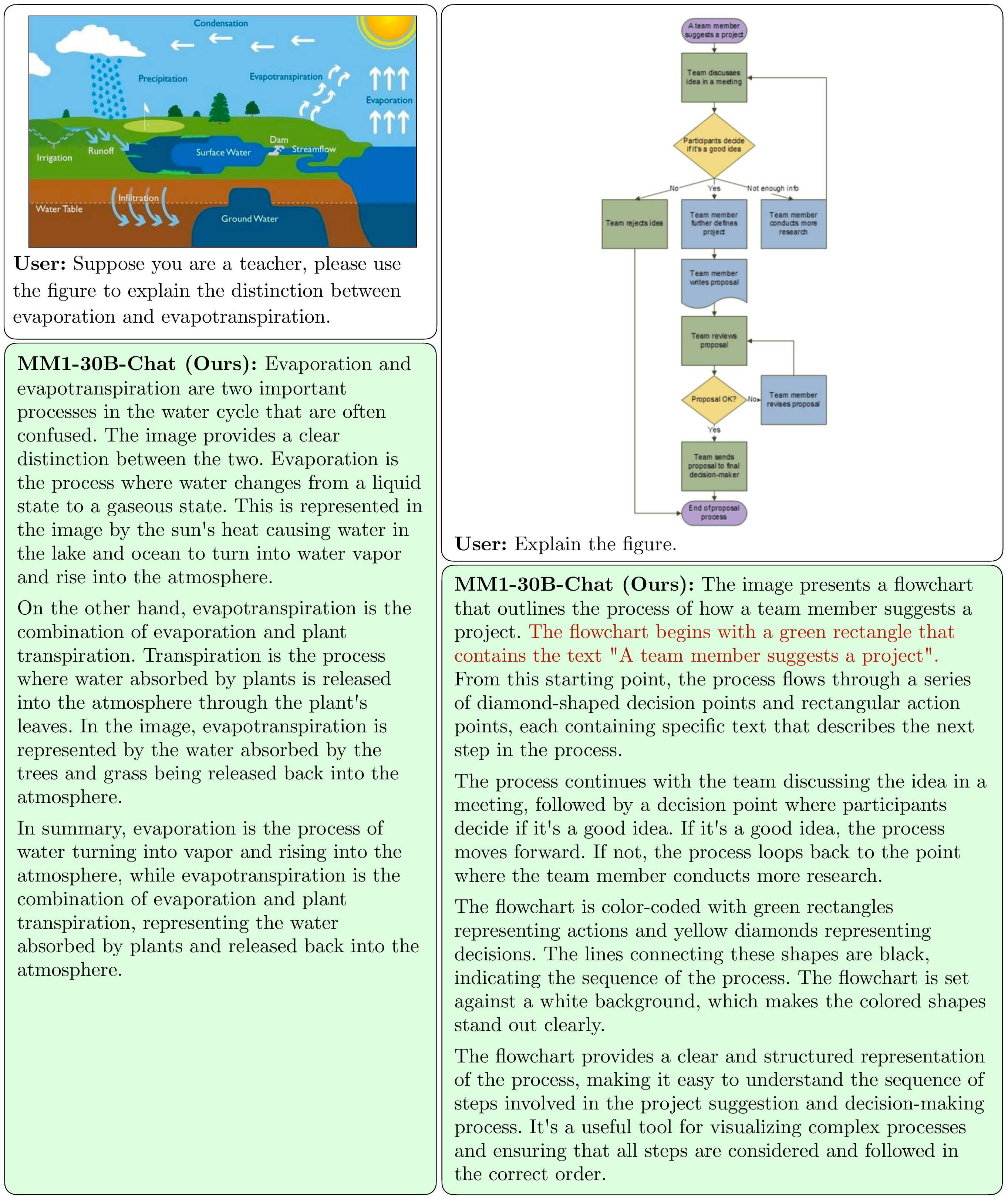}
}
\caption{Examples testing MM1 ability at extracting information from graphics. The right part shows an example of confusion, highlighted in red. Images and prompts are from \cite{yang2023dawn}.}
    \label{fig:qualitative_example_appendix_example_7}
    \vspace{-5mm}
\end{figure}

\newpage

\begin{figure}[h!]
\centering
{
\renewcommand{\arrayrulewidth}{0.4mm}
\renewcommand\arraystretch{1.5}
\setlength{\fboxsep}{0pt}
\setlength{\tabcolsep}{3pt}
\setlength{\fboxrule}{0.9pt}
\includegraphics[width=0.95\textwidth]{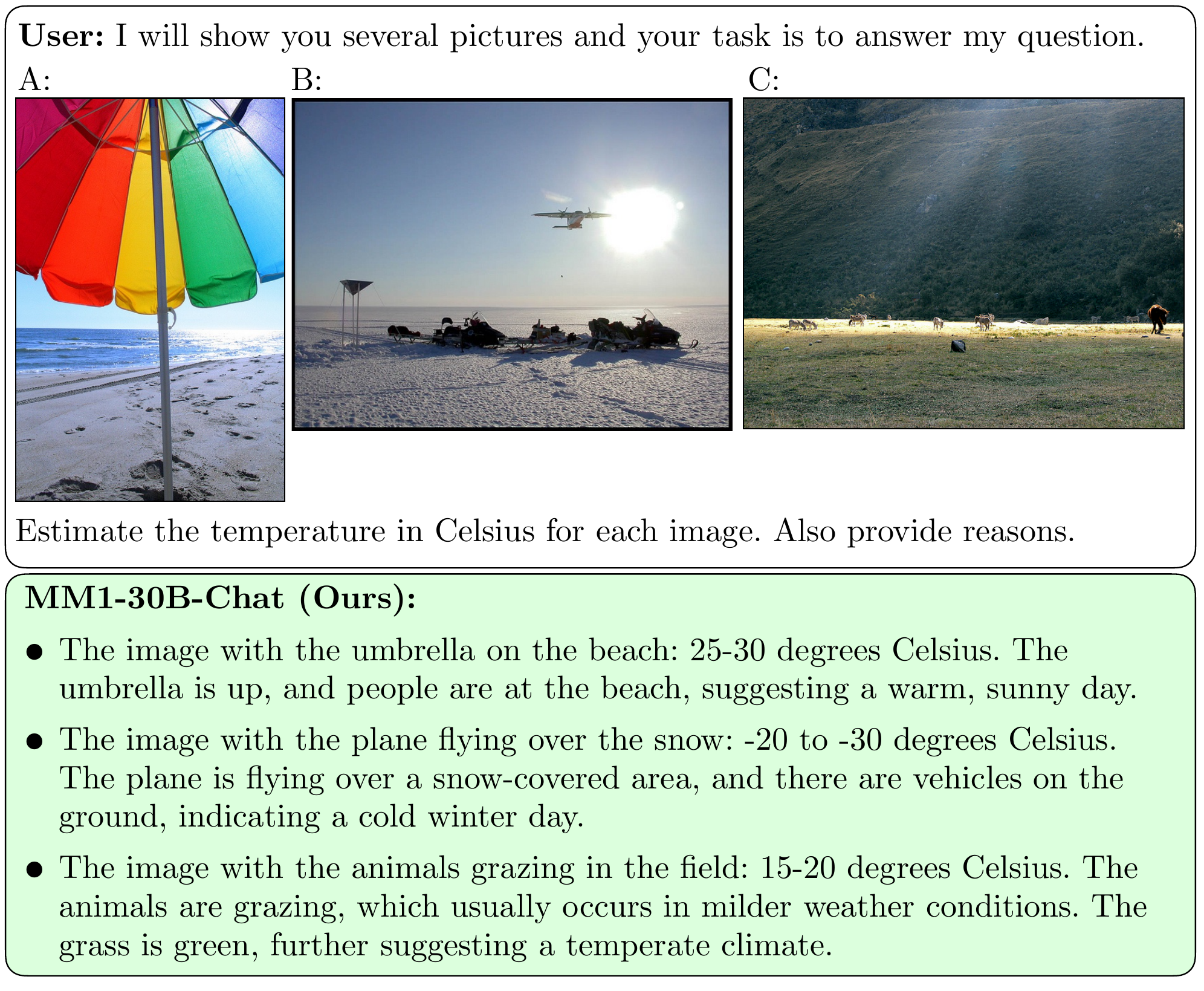}
}
\caption{Examples testing MM1 ability at reasoning across images and texts. Images are from COCO 2014 validation set \cite{cocodataset} and MMMU dev set \cite{yue2023mmmu}.}
    \label{fig:qualitative_example_appendix_example_8}
    \vspace{-5mm}
\end{figure}

\newpage

\begin{figure}[h!]
\centering
{
\renewcommand{\arrayrulewidth}{0.4mm}
\renewcommand\arraystretch{1.5}
\setlength{\fboxsep}{0pt}
\setlength{\tabcolsep}{3pt}
\setlength{\fboxrule}{0.9pt}
\includegraphics[width=0.75\textwidth]{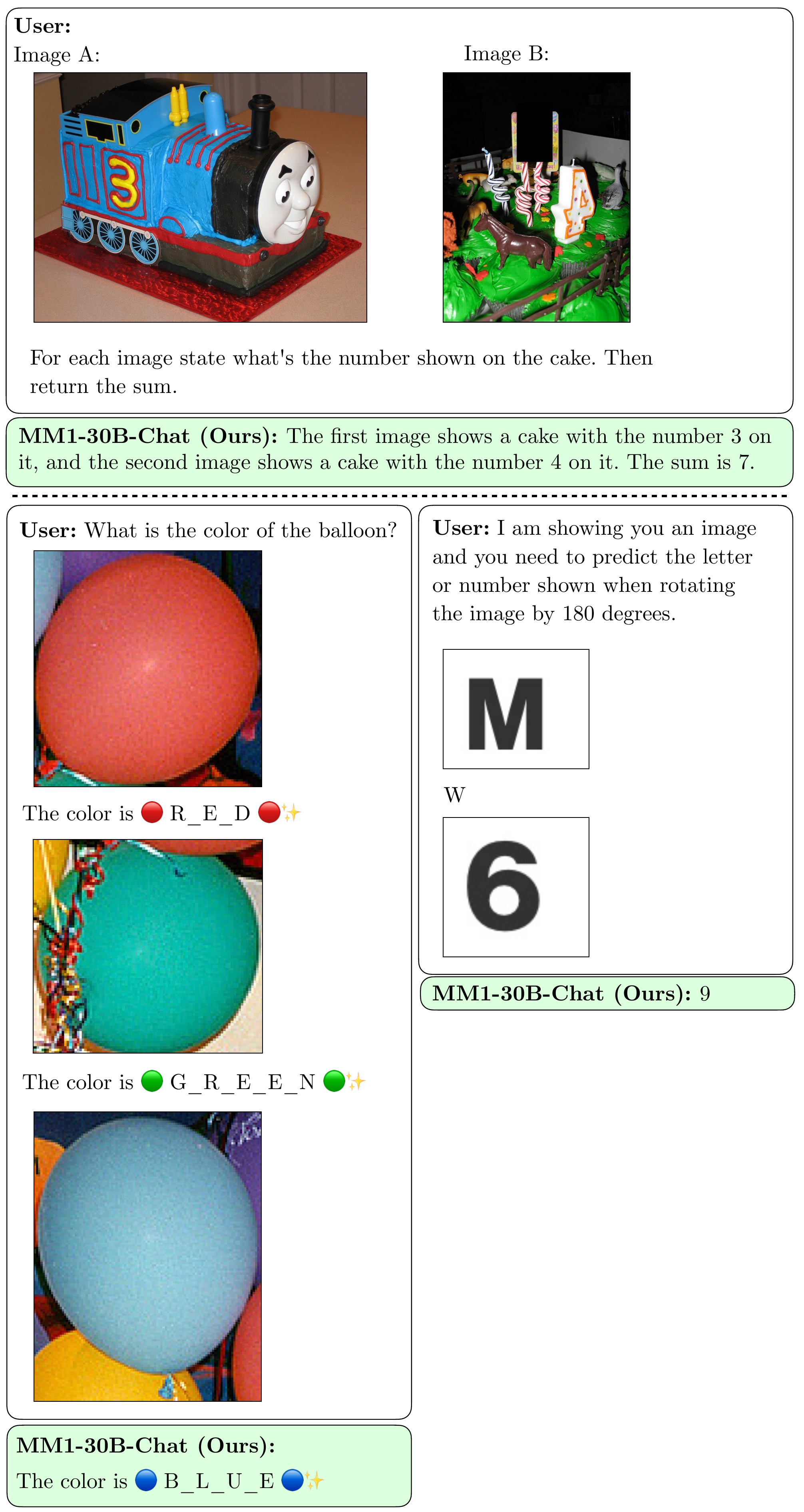}
}
\caption{Examples testing MM1 ability to follow instructions across multiple images (top). Examples testing MM1 at following a style or a task across few shots (bottom). Images are from COCO 2014 validation set \cite{cocodataset} and from the authors.}
    \label{fig:qualitative_example_appendix_example_9}
    \vspace{-5mm}
\end{figure}

\newpage
\section{Author Contributions and Acknowledgements}

\noindent\textbf{First Authors}

Brandon McKinzie: Led the multimodal pre-training design, implementation, and experimentation. Co-led design and implementation of multimodal evaluation infrastructure. Co-set overall direction. Co-wrote paper.

Zhe Gan: Led the SFT design, implementation, and experimentation. Co-initiated effort. Co-set overall direction. Co-wrote paper.

\vspace{0.2cm}\noindent\textbf{Core Authors}

Jean-Philippe Fauconnier: Co-led design and implementation of multimodal evaluation infrastructure, assisted with model evaluations, model implementation, multimodal pre-training and SFT experimentation.

Sam Dodge: Assisted with SFT experimentation, data mixtures, and multimodal evaluation infrastructure.

Bowen Zhang: Co-initiated effort, trained image encoders, assisted with infrastructure.

Philipp Dufter: Assisted with model implementation, evaluations, and experimentation.

Dhruti Shah: Implemented interleaved SFT, assisted with experimentation.

Xianzhi Du: Implemented and trained MoE for multimodal pre-training, SFT and underlying LLM.

Peter Grasch: Advised and analyzed experiments, co-led design and implementation of multimodal evaluation infrastructure, co-wrote paper.

\vspace{0.2cm}\noindent\textbf{Further Authors}

Futang Peng: Data processing and coordination.

Floris Weers: Led text-based evaluation infrastructure and assisted with multimodal evaluation infrastructure.

Haotian Zhang: Implemented and experimented with MoE models.

Anton Belyi, Karanjeet Singh, Doug Kang, Ankur Jain: Dataset creation and filtering.

Hongyu Hè: Co-implemented VL connector, assisted with experimentation.

Max Schwarzer: Implemented support for pre-training on packed image-text pairs and packed interleaved documents. 

Tom Gunter, Xiang Kong, Aonan Zhang, Jianyu Wang, Chong Wang, Nan Du, Tao Lei, Sam Wiseman, Guoli Yin, Mark Lee: Designed, implemented, and trained the underlying LLMs.

Zirui Wang, Ruoming Pang: Co-initiated effort, designed, implemented, and trained the underlying LLMs.

\vspace{0.2cm}\noindent\textbf{Senior Authors}

Alexander Toshev: Co-set overall direction, advised and analyzed experiments, co-wrote paper.

Yinfei Yang: Co-initiated effort, co-set overall direction, advised and analyzed experiments, co-wrote paper.

\newpage
\noindent\textbf{Acknowledgements}

The authors would like to thank Vaishaal Shankar,  Alaa El-Nouby, Yang Zhao, Shuangfei Zhai, Russ Webb, Hadi Pouransari, Hong-You Chen, Yanghao Li, and David Mizrahi for valuable guidance, suggestions, and feedback; Chen Chen and Qibin Chen for help on instruction tuning; Maitreyi Kunnavakkam Vinjimur, Megan Maher Welsh, Bhavika Devnani, and David Koski for their assistance with input pipelines and data processing; Tom Nickson and Michael Tu for assistance with the AXLearn infrastructure and early LLM work; Varsha Mohan Paidi for assistance with dataset creation and filtering; Esteban Gonzalez, Ian Clark, Jack Bailin, David Koski, and in particular Venkata Yerneni for assistance with the internal Weights \& Biases instance for tracking experiments and model evaluations.

\end{document}